\documentclass{article}
\PassOptionsToPackage{numbers, compress}{natbib}

\usepackage{amsmath}
\usepackage{multirow}
\usepackage{subcaption}
\usepackage{graphicx}
\usepackage[table]{xcolor}
\usepackage{placeins}
\usepackage{float}

\usepackage[preprint]{neurips_2026}

\usepackage[utf8]{inputenc}
\usepackage[T1]{fontenc}
\usepackage{hyperref}
\usepackage{url}
\usepackage{booktabs}
\usepackage{amsfonts}
\usepackage{nicefrac}
\usepackage{microtype}
\usepackage{xcolor}
\usepackage{newunicodechar}
\newunicodechar{，}{,}

\title{Probing Spectrum-Like Organization of States of Mind in Transformer Representation Spaces}

\author{%
  Sophie Zhao \\
  School of Computer Science\\
  Georgia Institute of Technology\\
  \texttt{sophie.zhao@gatech.edu,  sophie.m.zhao@gmail.com} \\
}

\begin{document}

\maketitle

\begin{abstract}
We investigate whether graded   states of mind form spectrum-like structure in transformer representation spaces. To do so, we construct a dataset of 636 short natural-language sentences annotated with both a continuous score from $-5$ to $5$ and one of seven ordered tiers, ranging from collapsed or scarcity-driven expressions to more coherent, reflective, and integrative ones. We evaluate five frozen transformer representations: four sentence-embedding models and one decoder-only residual-stream representation. Across all representations, simple probes reliably recover both the continuous score and the discrete tier labels, and permutation tests show that performance significantly exceeds shuffled-label baselines. Additional analyses reveal a consistent geometric pattern: UMAP projections show low-to-high organization, confusion matrices concentrate errors between neighboring tiers, and directional ablation identifies a prominent score-aligned component. These results suggest that transformer representations contain statistically significant, spectrum-like organization aligned with the annotated state-of-mind structure. The annotations are used only as an operational framework for representation analysis, not as a clinical or diagnostic measure.
\end{abstract}

\section{Introduction}

Transformer language models \citep{vaswani2017attention} represent text in high-dimensional spaces that encode rich semantic and contextual structure. Much of this structure is usually studied through downstream task performance, isolated probing tasks, or individual interpretable directions. In this work, we ask a broader geometric question: are graded states of mind recoverable, locally separable, and organized as a spectrum in transformer representation space?

We study this question using a dataset of 636 short natural-language sentences annotated with both a continuous score from $-5$ to $5$ and one of seven ordered state-of-mind tiers. Here, ``state of mind'' does not refer to a single discrete emotion, such as happiness, anger, or sadness. Instead, it refers to the overall psychological state or inner mode of organization expressed in a sentence: for example, whether the speaker expresses collapse, self-blame, scarcity, defensiveness, or control, as opposed to acceptance, reflection, growth, understanding, compassion, or integration. Accordingly, the tiers range from collapsed, scarcity-driven, or conflict-oriented expressions to more activated, reflective, and integrative ones. The annotation scheme draws conceptual inspiration from psychological and contemplative literature \citep{taoteching,varela1991embodied,jung1964man,wilber1977spectrum,cleary1993flower}, but is used here only as an operational framework for probing representation geometry, not as a clinical or diagnostic model.

We evaluate five frozen transformer representations: four sentence-embedding models (BGE, MPNet, MiniLM, and Qwen3-Embedding) and one decoder-only residual-stream representation (Qwen2.5-3B-Instruct layer 24). Across these representations, we combine probing, UMAP visualization, permutation testing, confusion-matrix analysis, and directional ablation to test whether the annotated structure is recoverable and geometrically organized.

Our results show that both the continuous score and discrete tier labels are significantly recoverable across all five representations. The observed organization is not explained by shuffled-label baselines or by a TF--IDF lexical baseline. UMAP projections reveal low-to-high gradients, classification errors are concentrated between neighboring tiers, and directional ablation identifies a prominent score-aligned component. Together, these findings suggest that transformer representation spaces contain statistically significant, spectrum-like organization aligned with our graded state-of-mind annotations.

\paragraph{Related Work.}
This work builds on probing, sentence representation geometry, and affective-language analysis. Prior work shows that contextual and sentence representations encode recoverable linguistic, semantic, and contextual properties \citep{ethayarajh2019contextual,reimers2019sentence,alain2016understanding,conneau2018probing,hewitt2019designing,belinkov2019analysis}. Other studies show that affective and psychological signals can be modeled in language, including emotion, valence, arousal, and related affective categories \citep{demszky2020goemotions,mohammad2018,reichman2026emotions,zhao2025hierarchical_emotion}. Recent representation-analysis work further suggests that abstract variables such as space, time, sentiment, truthfulness, persona, and emotion can align with low-dimensional or interpretable directions in language-model representations \citep{reichman2026emotions,zhao2025hierarchical_emotion, gurnee2024space,park2024linear,tigges2023linear,marks2024geometry,chen2025persona}. We differ by testing whether ordered state-of-mind annotations form a broader spectrum-like organization across fixed transformer representations, rather than focusing on a single attribute, emotion category, or contrastive direction.

\textbf{Contributions.}
We contribute: (1) a 636-sentence dataset with seven ordered state-of-mind tiers and continuous scores; (2) an evaluation across four sentence-embedding models and one decoder-only residual-stream representation; and (3) a geometric analysis showing that the annotated spectrum is statistically significant, locally ordered, and partially direction-aligned across transformer representation spaces.

\section{Methods}

\subsection{Dataset Construction}
\label{sec:dataset}

We construct a manually annotated dataset of 636 short natural-language sentences spanning a range of affective, reflective, and integrative expressions. Each sentence is assigned one of seven ordered state-of-mind tiers and a continuous score in $[-5,5]$.

\paragraph{State-of-mind tiers.}
The seven tiers are ordered from contracted to integrative states:
\textbf{Collapse} (depletion, despair, self-blame),
\textbf{Striving} (scarcity, fear, craving, insecurity),
\textbf{Conflict} (opposition, anger, dominance, control),
\textbf{Activation} (courage, resolve, acceptance, readiness),
\textbf{Growth} (learning, healing, transformation, contribution),
\textbf{Clarity} (reasoning, abstraction, understanding, coherence), and
\textbf{Unity} (compassion, peace, wholeness, non-dual integration).
The tiers are intended to capture modes of organization in language rather than emotional valence alone; the taxonomy is used only as an operational annotation framework.

\paragraph{Sentence construction and scores.}
Sentences were written to be short, mostly first-person, introspective, or experiential, with limited reliance on explicit emotion words as primary cues. Higher-tier sentences may use more reflective or contemplative phrasing. Each sentence belongs to exactly one tier. In addition, each sentence receives a continuous energy score from $-5$ to $+5$, representing a coarse ordinal position from contracted or depleted expression to more expansive or integrative expression. Scores may overlap across adjacent tiers, especially near boundaries.

\begin{table}[!htbp]
\centering
\caption{Representative sentences from each state-of-mind tier with illustrative scores.}
\label{tab:example-sentences}
\begin{tabular}{lp{10.2cm}c}
\toprule
Tier & Example Sentence & Score \\
\midrule
Collapse & ``Everything I do seems to make things worse, and I cannot see a way out.'' & $-4.5$ \\
Striving & ``I keep worrying that I am not doing enough, and that they might leave me.'' & $-2.9$ \\
Conflict & ``Why should I listen to people who are not at my level? Nobody knows better than me.'' & $-1.7$ \\
Activation & ``I can accept what is happening and bring myself back to center.'' & $0.0$ \\
Growth & ``I am learning from what happened and trying to respond differently this time.'' & $1.8$ \\
Clarity & ``Looking at the situation objectively helps me understand why it unfolded this way.'' & $3.0$ \\
Unity & ``I feel a quiet sense of connection and compassion, even in difficulty.'' & $4.2$ \\
\bottomrule
\end{tabular}
\end{table}

\paragraph{Dataset intent.}
The labels and scores are not treated as precise psychological measurements. They are coarse semantic annotations designed for controlled representation-geometry experiments. The dataset is not intended as a clinical, diagnostic, or psychological assessment framework. Additional examples and annotation notes are provided in Appendix~\ref{app:more_sentence_example}.

\subsection{Transformer Representations and Probing}
\label{sec:embeddings}

We evaluate five frozen transformer representations. Four are sentence-embedding models:
\texttt{BAAI/bge-large-en-v1.5} \citep{xiao2024cpack},
\texttt{all-mpnet-base-v2} \citep{song2020mpnet},
\texttt{all-MiniLM-L6-v2} \citep{wang2020minilm}, and
\texttt{Qwen3-Embedding-0.6B} \citep{zhang2025qwen3embedding}.
The fifth is a decoder-only residual-stream representation: the layer-24 final-token hidden state of \texttt{Qwen2.5-3B-Instruct}. We use layer 24 because a preliminary sweep over candidate residual-stream layers found it to best reflect the annotated structure, using diagnostics for score decodability, tier separability, local tier coherence, and cluster separation. All representations are extracted without fine-tuning.

To test whether the annotations are recoverable from these representations, we train simple probes on fixed vectors. For the continuous score, we use Ridge regression and a shallow MLP regressor, reporting $R^2$ and MSE. For the seven ordered tiers, we use multinomial logistic regression and a shallow MLP classifier, reporting accuracy and weighted F1. Linear probes provide a low-capacity test of approximate separability, while shallow nonlinear probes test whether modest nonlinear boundaries improve recovery. Results are averaged over 20 random 80/20 train--test splits.

We also inspect confusion matrices from a representative split to test whether errors are local along the tier order. Misclassifications concentrated between neighboring tiers, such as Activation--Growth or Growth--Clarity, would indicate graded organization rather than arbitrary class separability.

\subsection{UMAP Visualization}
\label{sec:umap}

To qualitatively inspect geometry, we apply UMAP \citep{mcinnes2018umap} to each representation set and color points by continuous score in $[-5,5]$. We report 3D UMAP plots in the main paper and 2D plots in Appendix~\ref{app:2D_umap_all}.

\subsection{Permutation Tests}
\label{sec:permutation}

To test whether probe performance reflects alignment between representations and annotations rather than chance correlations, we use nonparametric label-permutation tests \citep{good2013permutation}. We evaluate two null hypotheses: continuous scores are independent of representations, and tier labels are independent of representations. Under each null, labels are randomly permuted while representations $X$ are fixed, and the same probing protocol is rerun using 20 fixed 80/20 train--test splits. The test statistics are mean Ridge $R^2$ for score regression and mean weighted F1 for tier classification. Monte Carlo details are provided in Appendix~\ref{app:montecarlo}.

\subsection{Directional Ablation}
\label{sec:directional_ablation}

To test whether score-related information is concentrated along a specific direction, we fit a Ridge probe from representations $X \in \mathbb{R}^{N \times d}$ to scores $y$ and interpret its weight vector $w$ as a candidate score-aligned direction. After normalizing $\hat{w}=w/\|w\|_2$, we remove each vector's projection onto this direction:
\begin{equation}
    x' = x - (x \cdot \hat{w})\hat{w}.
\end{equation}
The ablated representations are then evaluated using the same probing protocol.

We also measure the variance share captured by the learned direction:
\begin{equation}
    \mathrm{Ratio} =
    \frac{\mathrm{Var}(X_c \hat{w})}
    {\mathrm{trace}(\mathrm{Cov}(X_c))},
\end{equation}
where $X_c$ denotes mean-centered representations. As controls, we repeat the ablation using random directions and directions learned from permuted score labels.

\subsection{TF--IDF Baseline}
\label{sec:tfidf_method}

We include a TF--IDF unigram--bigram baseline evaluated with the same probing, confusion-matrix, and UMAP analyses. This tests whether the observed structure can be explained by lexical statistics alone.

\section{Results}

\subsection{Decodability of Continuous Energy Scores}
\label{sec:results_energy}

We first evaluate whether the continuous energy scores assigned to sentences are recoverable from fixed transformer representations. This analysis tests whether representation geometry preserves graded structure aligned with a low-to-high state-of-mind spectrum.

Across all evaluated representations, energy scores are strongly decodable. As shown in Table~\ref{tab:energy_regression}, regression performance is well above chance, capturing a substantial proportion of the annotated ordinal signal on held-out data. BGE, Qwen3-Embedding, and Qwen2.5 L24 all achieve strong score decodability, with Ridge $R^2$ values around $0.76$--$0.79$ and MLP $R^2$ values around $0.78$--$0.82$. MPNet is also strong, while MiniLM remains lower but substantially above chance.

\begin{table}[!htbp]
\centering
\small
\caption{Energy regression probe performance averaged over 20 train--test splits. Qwen2.5 L24 denotes layer-24 residual-stream activations rather than a sentence-embedding model.}
\label{tab:energy_regression}
\begin{tabular}{lcccc}
\toprule
Representation & Ridge $R^2$ $\uparrow$ & Ridge MSE $\downarrow$ & MLP $R^2$ $\uparrow$ & MLP MSE $\downarrow$ \\
\midrule
BGE-large-en-v1.5 & 0.791 & 1.762 & 0.820 & 1.512 \\
MPNet-base-v2 & 0.743 & 2.165 & 0.756 & 2.049 \\
MiniLM-L6-v2 & 0.634 & 3.085 & 0.652 & 2.936 \\
Qwen3-Embedding-0.6B & 0.758 & 2.043 & 0.785 & 1.813 \\
Qwen2.5-3B-Instruct L24 & 0.781 & 1.846 & 0.822 & 1.494 \\
\bottomrule
\end{tabular}
\end{table}

Decodability improves modestly when moving from linear to shallow nonlinear regression. Across all representations, nonlinear probes achieve slightly higher $R^2$ and lower mean squared error than their linear counterparts.

However, this improvement should be interpreted cautiously. While it may indicate that some energy-score-related variation is not perfectly captured by a single linear direction, it may also reflect the coarse and manually assigned nature of the annotation scheme. Because energy scores are intended as approximate ordinal signals rather than precise interval-scaled measurements, shallow nonlinear models may partially compensate for annotation noise or boundary ambiguity rather than revealing intrinsic nonlinear structure in the representation space.

We also observe model-dependent differences. BGE achieves the strongest linear score regression among the transformer representations, while Qwen3-Embedding and Qwen2.5 L24 also show strong continuous-score decodability. This suggests that the score-aligned structure is not specific to one model family.

Overall, these results provide strong evidence that continuous energy scores are consistently recoverable from transformer representation spaces, suggesting that graded energy-related structure is reflected in their geometry.

\subsection{Decodability of State-of-Mind Tiers}
\label{sec:results_tiers}

We next examine whether the discrete state-of-mind tiers assigned to sentences are recoverable from transformer representations. Unlike the continuous energy scores in Section~\ref{sec:results_energy}, tier labels represent coarser categorical stages along the same underlying spectrum.

Across all representations, tier labels are substantially decodable, with classification performance well above chance. As summarized in Table~\ref{tab:tier_classification}, weighted F1-scores range from approximately $0.57$ to $0.71$ for linear probes and from approximately $0.63$ to $0.73$ for shallow nonlinear probes, indicating that the representation space preserves meaningful separation among the seven tiers.

\begin{table}[!htbp]
\centering
\small
\caption{Tier classification probe performance averaged over 20 train--test splits. Qwen2.5 L24 denotes layer-24 residual-stream activations rather than a sentence-embedding model.}
\label{tab:tier_classification}
\begin{tabular}{lcccc}
\toprule
 & \multicolumn{2}{c}{Logistic Regression} & \multicolumn{2}{c}{MLP (128,64)} \\
\cmidrule(lr){2-3}\cmidrule(lr){4-5}
Representation & Accuracy $\uparrow$ & Weighted F1 $\uparrow$ & Accuracy $\uparrow$ & Weighted F1 $\uparrow$ \\
\midrule
BGE-large-en-v1.5 & 0.729 & 0.712 & 0.727 & 0.715 \\
MPNet-base-v2 & 0.718 & 0.705 & 0.713 & 0.705 \\
MiniLM-L6-v2 & 0.664 & 0.652 & 0.647 & 0.635 \\
Qwen3-Embedding-0.6B & 0.723 & 0.704 & 0.736 & 0.727 \\
Qwen2.5-3B-Instruct L24 & 0.615 & 0.568 & 0.689 & 0.675 \\
\bottomrule
\end{tabular}
\end{table}

Consistent with the regression results, performance varies across representations. 
Among the sentence-embedding models, BGE achieves the strongest linear tier classification, 
while MPNet and Qwen3-Embedding perform similarly. Qwen3-Embedding achieves the strongest 
shallow nonlinear tier F1, suggesting that its tier-related structure may benefit from modest 
nonlinear decision boundaries. Qwen2.5 L24 shows strong score regression but weaker linear tier 
classification, suggesting that its residual-stream representation may encode the continuous 
spectrum more clearly than the discrete tier boundaries.

\begin{figure}[!htbp]
\centering
\includegraphics[width=0.45\linewidth]{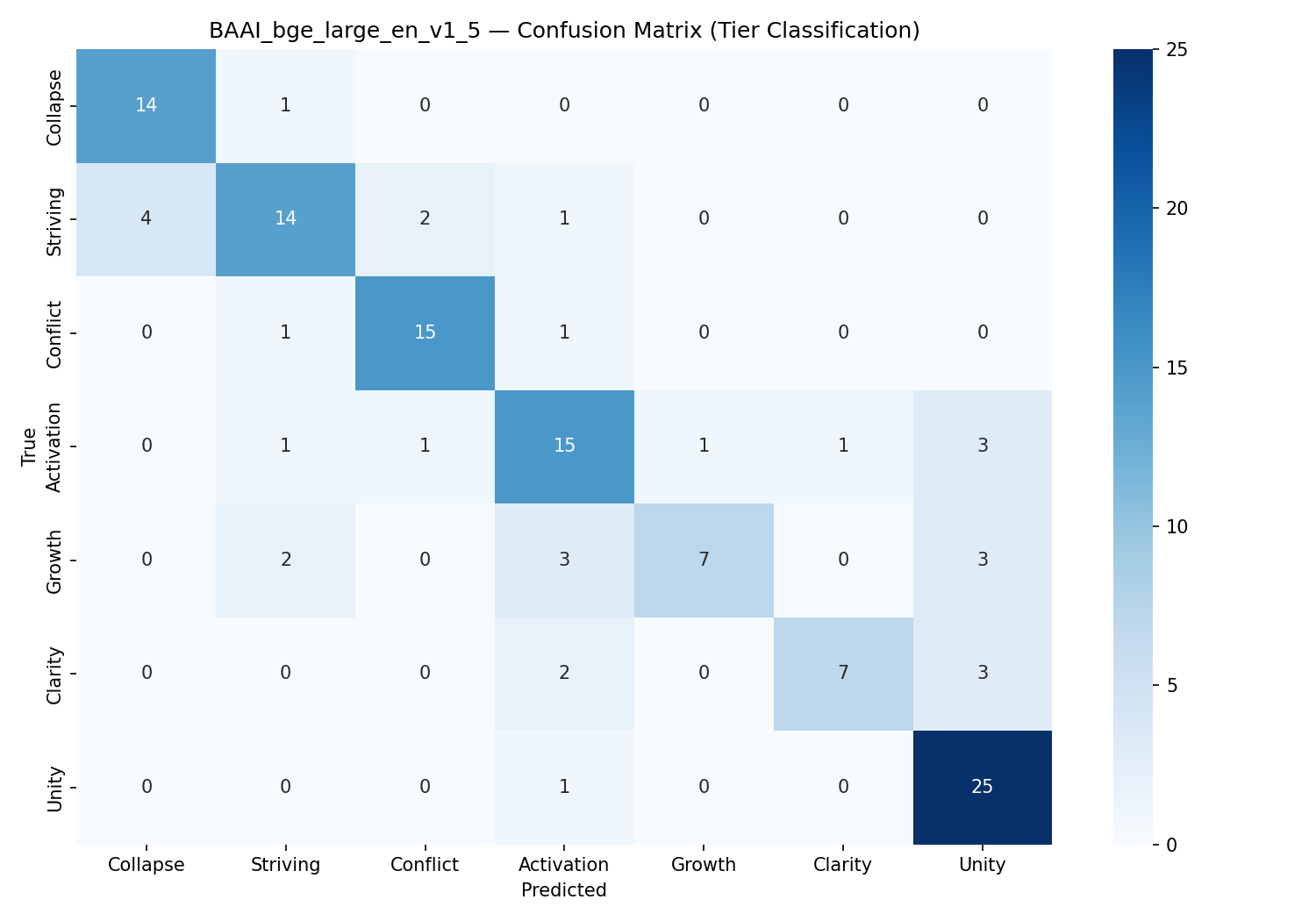}
\caption{Confusion matrix for linear tier classification using BAAI-bge-large-en-v1.5 representations.}
\label{fig:confusion_m1}
\end{figure}

To better understand the nature of classification errors, we inspect confusion matrices for a representative train--test split. Figure~\ref{fig:confusion_m1} shows the confusion matrix for tier classification using BAAI/bge-large-en-v1.5 embeddings. Misclassifications are concentrated between adjacent or nearby tiers, such as Growth $\leftrightarrow$ Activation and Striving $\leftrightarrow$ Collapse, while confusions between distant tiers, such as Striving $\leftrightarrow$ Clarity, are rare. Similar patterns are observed for other representations (Appendix~\ref{app:confusion_mat_all}).

Taken together, these results indicate that state-of-mind tiers are not only decodable, but are organized in an ordered fashion within the representation space. The concentration of errors between neighboring tiers suggests that representations encode a graded structure consistent with the proposed tiers.

\subsection{Qualitative Structure in Representation Space via UMAP}
\label{sec:results_umap}

\begin{figure}[!htbp]
\centering
\includegraphics[width=\linewidth]{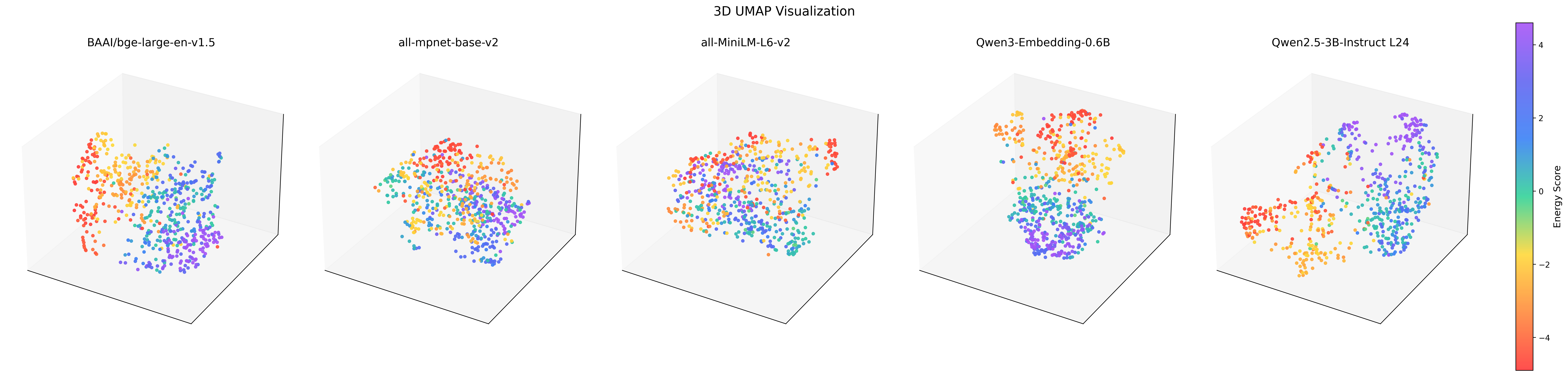}
\caption{3D UMAP visualization of five transformer representations colored by energy scores.}
\label{fig:umap_3d}
\end{figure}

To complement quantitative probing results, we examine the geometric organization of sentences in representation space using UMAP visualizations \citep{mcinnes2018umap} colored by continuous energy scores (Figure~\ref{fig:umap_3d}).

Across all five representations, UMAP reveals a clear low-to-high energy gradient rather than random mixing. Sentences annotated with lower energy scores, such as Collapse and Striving, tend to occupy contiguous regions, while higher-energy sentences, such as Clarity and Unity, occupy distinct regions of the representation space.

Model-dependent differences are apparent. BAAI/bge-large-en-v1.5 and the two Qwen representations exhibit particularly coherent low-to-high organization, with relatively smooth transitions from lower-score to higher-score regions. MPNet-base-v2 shows a similar global gradient but with increased overlap between adjacent score levels. MiniLM-L6-v2 displays greater dispersion and mixing, consistent with its weaker regression and classification performance. These differences suggest that the clarity of the visual geometry depends not only on model size, but also on training objective and representation type.

Importantly, misalignments observed in the confusion matrices, primarily between adjacent tiers, are reflected in UMAP by local overlaps rather than long-range mixing. Distant tiers, such as Collapse versus Unity, rarely occupy the same regions, suggesting that representation geometry preserves a coarse order even when fine-grained boundaries are ambiguous.

\subsection{Statistical Significance via Permutation Tests}
\label{sec:results_permutation}

To assess whether the observed probe performance reflects a genuine alignment between representation geometry and the annotated state-of-mind attributes, rather than spurious correlations, we evaluate statistical significance using nonparametric permutation tests (Section~\ref{sec:permutation}).

\begin{table}[!htbp]
\centering
\footnotesize
\caption{Permutation test results across five representations ($N_{\mathrm{perm}}=200$). All observed probe scores exceed shuffled-label baselines, achieving the minimum attainable smoothed one-sided $p$-value, $1/(200+1)=0.00498$.}
\label{tab:permutation_all_models}
\begin{tabular}{lcccc}

\toprule
Representation & Ridge $R^2$ & $p$-value & Tier F1 & $p$-value \\
\midrule
BGE-large-en-v1.5 & 0.791 & 0.00498 & 0.715 & 0.00498 \\
MPNet-base-v2 & 0.740 & 0.00498 & 0.700 & 0.00498 \\
MiniLM-L6-v2 & 0.632 & 0.00498 & 0.653 & 0.00498 \\
Qwen3-Embedding-0.6B & 0.762 & 0.00498 & 0.699 & 0.00498 \\
Qwen2.5-3B-Instruct L24 & 0.778 & 0.00498 & 0.568 & 0.00498 \\
\bottomrule
\end{tabular}
\end{table}

Across representations, observed probe performance averaged across repeated 80/20 train--test splits lies far outside the permutation null distributions. Under the score permutation null, the empirical distribution of mean Ridge $R^2$ values remains well below the observed statistics, yielding the minimum attainable smoothed one-sided permutation value, $p=1/(200+1)=0.00498$, for every representation. A similar pattern is observed under the tier permutation null, where weighted F1 scores consistently exceed shuffled-label baselines for all five representations.

\begin{figure}[!htbp]
\centering
\begin{subfigure}[t]{0.48\linewidth}
    \centering
    \includegraphics[width=\linewidth]{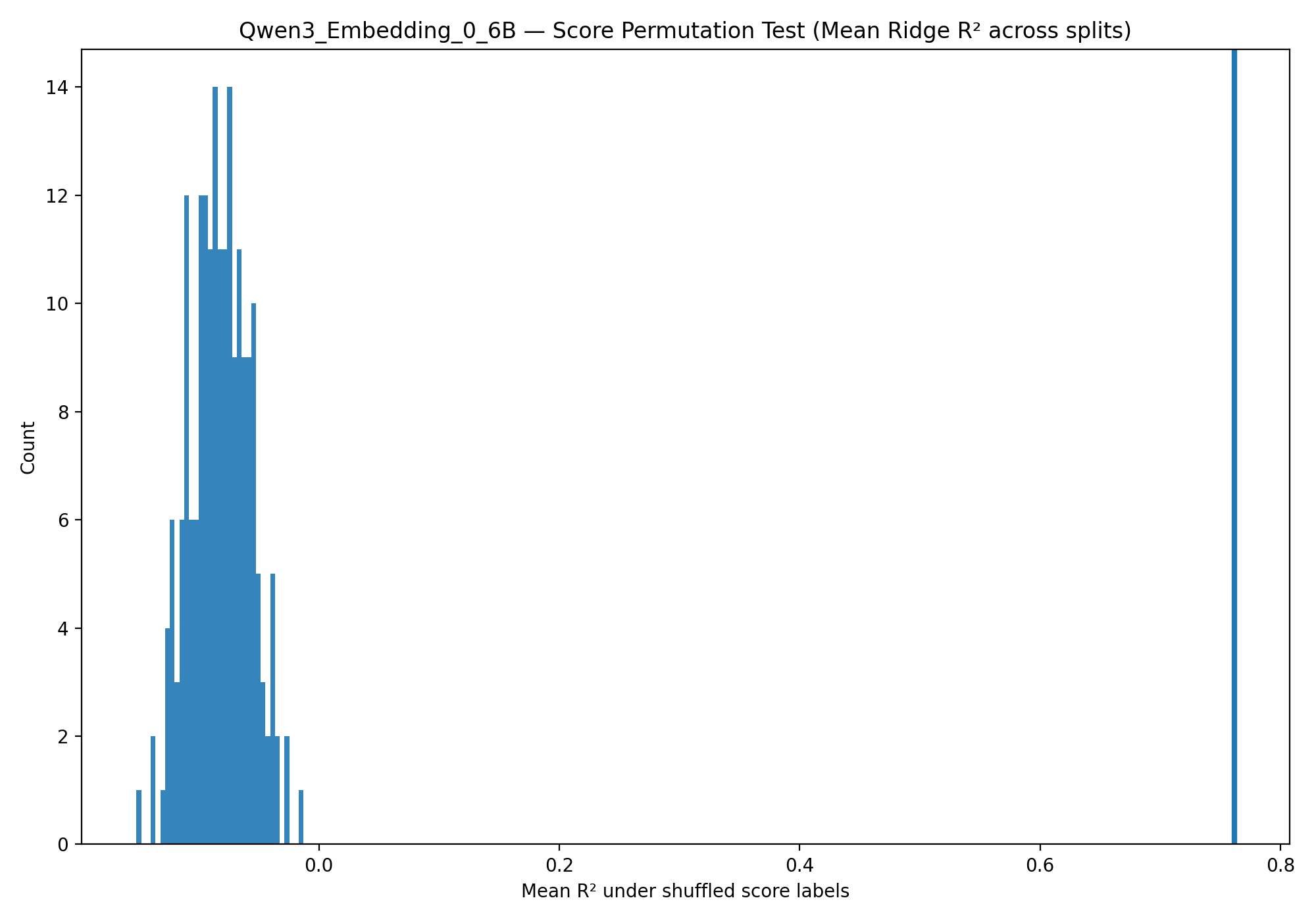}
    \caption{Score permutation test (Ridge $R^2$)}
    \label{fig:perm_score}
\end{subfigure}
\hfill
\begin{subfigure}[t]{0.48\linewidth}
    \centering
    \includegraphics[width=\linewidth]{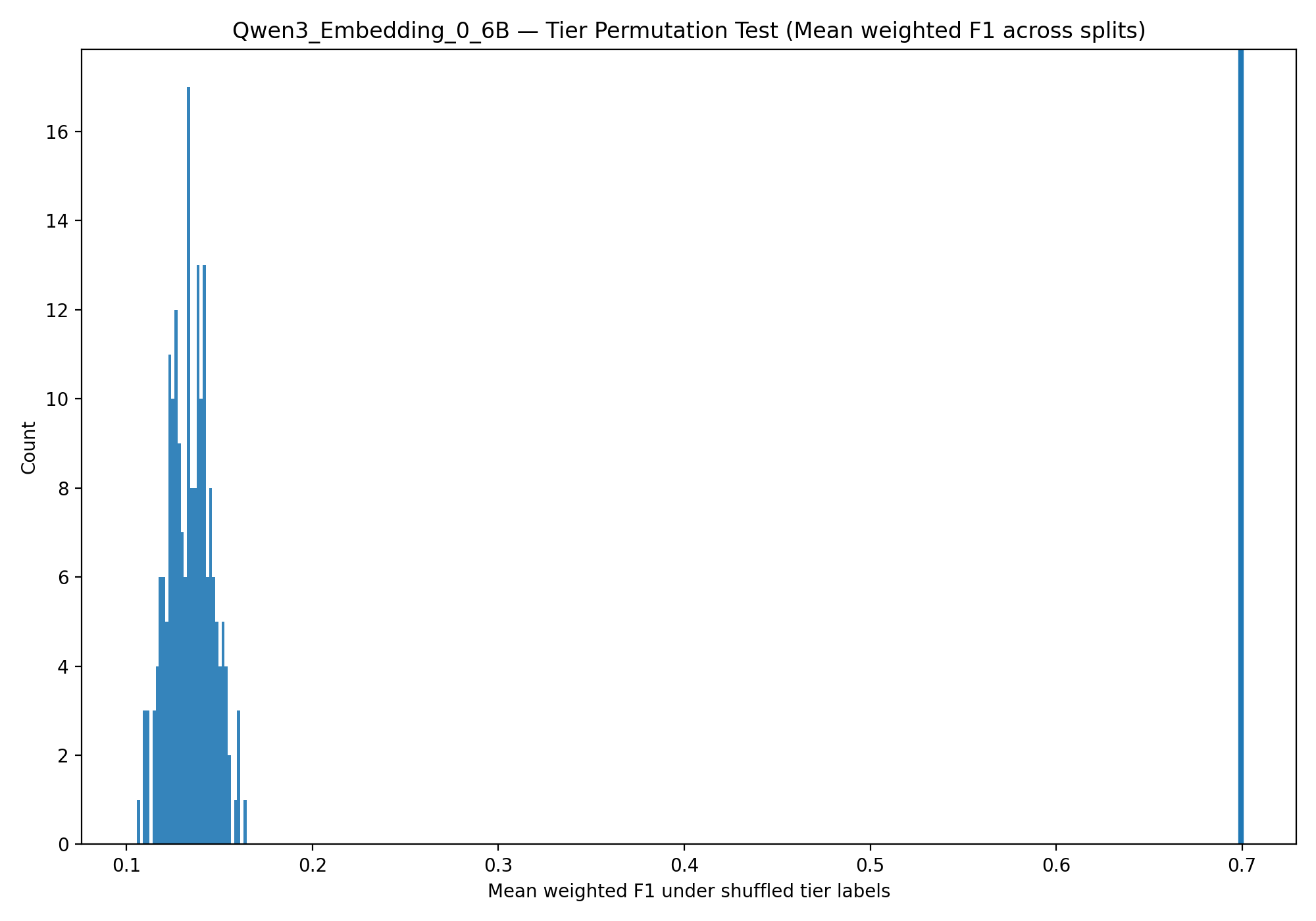}
    \caption{Tier permutation test (Weighted F1)}
    \label{fig:perm_tier}
\end{subfigure}
\caption{Permutation test results for Qwen3-Embedding-0.6B. Vertical lines indicate observed probe performance.}
\label{fig:perm_test}
\end{figure}

As shown in Figure~\ref{fig:perm_test}, empirical null distributions remain concentrated near chance performance, while observed probe statistics lie in the extreme upper tail.

Although absolute scores vary across representations, observed performance for all five remains well separated from permutation null distributions. Histogram visualizations for additional models are provided in Appendix~\ref{app:permutation_test_extra_his}.

Together, these findings indicate that transformer representations encode information aligned with both continuous energy scores and discrete state-of-mind tiers. The consistency of permutation significance across representations suggests that the observed separability reflects robust geometric organization rather than model-specific effects.

\subsection{Directional Ablation}
\label{sec:directional_ablation_results}

We next evaluate how removing the learned energy-aligned direction affects
downstream probing performance. This analysis tests whether energy-related
information is concentrated along a specific geometric component in embedding
space.

Table~\ref{tab:energy_regression_ablation_compare} compares energy regression performance before and after
directional ablation. Removing the learned direction substantially reduces regression
performance across all models. For example, Ridge $R^2$ drops from 0.791 to 0.499 for
\texttt{bge-large-en-v1.5}, indicating that a significant portion of the energy-related signal
is concentrated along a specific geometric component. Nonlinear probes exhibit a similar decline.

\begin{table}[!htbp]
\centering
\caption{Energy regression probe performance under directional ablation.
Normal denotes the original representations, ablated denotes projection
removal along the learned energy direction, and Perm-label control denotes
removal of a direction learned from permuted labels. Results are averaged over
20 train--test splits.}
\label{tab:energy_regression_ablation_compare}
\footnotesize
\begin{tabular}{llcccc}
\hline
Model & Condition & Ridge $R^2$ $\uparrow$ & Ridge MSE $\downarrow$ & MLP $R^2$ $\uparrow$ & MLP MSE $\downarrow$ \\
\hline

BGE-large-en-v1.5
& Normal          & 0.791 & 1.762 & 0.820 & 1.512 \\
& Ablated  & 0.499 & 4.213 & 0.550 & 3.778 \\
& Perm-label ctrl & 0.789 & 1.772 & 0.817 & 1.542 \\
\hline

MPNet-base-v2
& Normal  & 0.743 & 2.165 & 0.756 & 2.049 \\
& Ablated & 0.400 & 5.041 & 0.489 & 4.287 \\
& Perm-label ctrl & 0.740 & 2.188 & 0.754 & 2.064 \\
\hline

MiniLM-L6-v2
& Normal          & 0.634 & 3.085 & 0.652 & 2.936 \\
& Ablated & 0.262 & 6.199 & 0.352 & 5.449 \\
& Perm-label ctrl & 0.632 & 3.092 & 0.652 & 2.928 \\
\hline

Qwen3-Embedding-0.6B
& Normal          & 0.758 & 2.043 & 0.785 & 1.813 \\
& Ablated  & 0.439 & 4.710 & 0.521 & 4.018 \\
& Perm-label ctrl & 0.761 & 2.008 & 0.789 & 1.768 \\
\hline

Qwen2.5-3B-Instruct L24
& Normal          & 0.781 & 1.846 & 0.822 & 1.494 \\
& Ablated & 0.273 & 6.113 & 0.282 & 6.031 \\
& Perm-label ctrl & 0.776 & 1.880 & 0.818 & 1.519 \\
\hline

\end{tabular}
\end{table}

Additional analyses are reported in Appendix~\ref{app:directional_ablation}. The learned direction accounts for a measurable share of total variance, ranging from 1.63\% for MiniLM to 4.82\% for Qwen2.5 L24. Directions learned from permuted labels capture substantially less variance, and random directions capture still less on average. These findings indicate that the annotation-aligned direction is geometrically prominent across both sentence embedding spaces and decoder residual-stream activations.

\subsection{TF--IDF Baseline}
\label{sec:tfidf_results}

\begin{table}[!htbp]
\centering
\small
\caption{Comparison between transformer representations and a TF--IDF lexical baseline. Results are averaged over 20 train--test splits.}
\label{tab:tfidf_vs_embeddings}
\begin{tabular}{lcccc}
\toprule
Representation & Ridge $R^2$ $\uparrow$ & Ridge MSE $\downarrow$ & Tier Accuracy $\uparrow$ & Weighted F1 $\uparrow$ \\
\midrule
BGE-large-en-v1.5 & 0.791 & 1.762 & 0.729 & 0.712 \\
MPNet-base-v2 & 0.743 & 2.165 & 0.718 & 0.705 \\
MiniLM-L6-v2 & 0.634 & 3.085 & 0.664 & 0.652 \\
Qwen3-Embedding-0.6B & 0.758 & 2.043 & 0.723 & 0.704 \\
Qwen2.5-3B-Instruct L24 & 0.781 & 1.846 & 0.615 & 0.568 \\
\midrule
TF-IDF (1--2gram) & 0.444 & 5.296 & 0.473 & 0.421 \\
\bottomrule
\end{tabular}
\end{table}

As shown in Table~\ref{tab:tfidf_vs_embeddings}, the TF--IDF baseline performs substantially worse than transformer representations across both regression and classification tasks, suggesting that the observed structure cannot be accounted for by lexical statistics alone. Additional TF--IDF analyses and visualizations are provided in Appendix~\ref{app:tfidf}.

\section{Discussion}

\subsection{What Is Encoded in the Representation Geometry?}

Across five transformer representations, both continuous scores and discrete state-of-mind tiers are reliably decodable from fixed representations, suggesting a graded geometric organization aligned with the annotations. For score prediction, linear probes recover a substantial portion of the signal, while shallow nonlinear probes provide moderate additional gains. This pattern suggests that the score-related structure is strongly aligned with broad geometric directions, but is not purely one-dimensional. For tier classification, linear and shallow nonlinear probes perform comparably in most sentence-embedding models, indicating that tier boundaries are partly accessible through simple decision surfaces.

The observed organization is unlikely to be explained by lexical statistics alone. A TF--IDF baseline performs substantially worse than transformer representations across regression and classification tasks, while permutation tests show that probe performance lies far outside shuffled-label null distributions. Directional ablation provides further evidence: removing the learned score-aligned direction substantially reduces score prediction despite removing only a small fraction of total representation variance. Together, these results suggest that the annotated spectrum is not uniformly distributed across dimensions, but is partially concentrated along geometrically meaningful directions.

\subsection{Global Ordering and Local Ambiguity}

The analyses suggest that representation space preserves a coarse global ordering while allowing local ambiguity between neighboring tiers. Confusion matrices show that misclassifications primarily occur between adjacent or nearby tiers, such as Growth versus Activation or Striving versus Collapse, whereas distant confusions such as Collapse versus Unity are rare. UMAP visualizations show a similar pattern, with gradual low-to-high organization and local overlap among nearby score regions.

This behavior is consistent with a graded spectrum rather than sharply separated classes. Because state-of-mind language is context-dependent and semantically continuous, adjacent tiers are expected to overlap. The discrete labels can therefore be interpreted as approximate regions along a shared representational continuum rather than as rigid categories.

\subsection{Model Differences and Representation Type}

Comparisons across the five representations reveal systematic differences in how the annotated structure appears. Among the sentence-embedding models, BGE achieves the strongest linear score regression, suggesting that its embedding space most clearly aligns with the continuous score axis. MPNet performs slightly below BGE but remains strong across both regression and tier classification, indicating a similar but somewhat less pronounced organization. MiniLM shows the weakest performance among the embedding models, although its results remain statistically significant, consistent with a smaller representation model preserving the same structure with lower resolution.

Qwen3-Embedding-0.6B performs competitively with BGE and MPNet and achieves the strongest shallow nonlinear tier classification, suggesting that its representation space captures tier-related structure that may benefit from modest nonlinear decision boundaries. In contrast, the Qwen2.5-3B-Instruct layer-24 residual representation shows a particularly prominent score-aligned direction despite weaker linear tier separability. This suggests that decoder-only residual streams may encode the continuous score strongly while organizing discrete tier boundaries differently from dedicated sentence-embedding models.

These differences indicate that graded state of mind structure is influenced by architecture, training objective, and representation type. At the same time, significant decodability and permutation-test significance across all five representations suggest that the observed organization is not specific to a single model family.

\paragraph{Broader Impact.}
Our findings suggest that representation spaces may contain localized regions associated with psychologically sensitive or unstable states. While such structure could support beneficial applications such as interpretability or safety analysis, misuse may arise if representation geometry is deliberately leveraged to steer model behavior toward unstable regions. The present work is intended as a representation-analysis study rather than a clinical or diagnostic framework, and any downstream use involving psychological or safety-sensitive language should require careful validation, safeguards, and human oversight.

\paragraph{Limitations.}
This study is limited by dataset scale and scope. The dataset contains 636 short English sentences designed for controlled representation-geometry experiments, rather than a large naturalistic corpus. Thus, the findings should be interpreted as evidence under this annotation framework, not as a general claim about all state-of-mind language. Future work should test larger corpora, naturalistic dialogue, additional languages, and alternative taxonomies.

\paragraph{Conclusion.}
This work examines how human-defined state-of-mind annotations relate to representation geometry in transformer representations. Results show that both continuous scores and discrete tier labels correspond to approximately linear organization within representation space across five representations, including four sentence-embedding models and one decoder-only residual-stream representation. Instead of being evenly distributed across dimensions, annotated signals appear partially concentrated along specific geometric directions that remain consistent across multiple transformer representation spaces. Linear probes capture a large portion of this structure, while shallow nonlinear probes provide modest improvements, suggesting that the dominant organization is accessible through relatively simple geometric features.

From a geometric perspective, one consistent pattern across analyses is that sentences associated with similar state-of-mind annotations tend to occupy nearby regions in representation space. UMAP projections, probe performance, permutation tests, and confusion matrix structures collectively suggest that local neighborhoods preserve similarity in annotated attributes, while transitions between adjacent tiers occur gradually rather than through abrupt boundaries.

These findings suggest several directions for future research. The localization of state-of-mind-related language within specific regions of representation space may support representation-level safety analysis and interpretability, where representation geometry serves as a diagnostic signal for analyzing model behavior. For example, localized regions could help study shifts toward higher-risk or coercive linguistic patterns without relying solely on keyword-based filtering. More broadly, the observed geometric organization motivates exploration of representation-space monitoring or steering approaches, in which generation trajectories are analyzed relative to geometric regions associated with different attributes. Future work may further evaluate robustness across datasets, languages, and annotation paradigms, explore alternative state-of-mind taxonomies, and investigate whether similar graded organization emerges within intermediate transformer layers or during generation dynamics.

{\small
\bibliographystyle{unsrtnat}
\bibliography{main}
}

\appendix

\section{Additional Example Sentences per Tier}
\label{app:more_sentence_example}

This appendix provides additional illustrative example sentences for each tier to support qualitative interpretability. These examples are intended to help readers develop an intuitive sense of the state-of-mind patterns indexed by each tier, independent of access to the full annotated dataset.

\paragraph{Tier 1: Highly Contracted / Self-Destructive, Apathy, Despair}
\begin{itemize}
    \item ``The feeling was like standing in a tunnel with no light at either end.''
    \item ``I can’t stop blaming myself for what happened.''
    \item ``I don't care anymore - well, nobody cares about me anyways.''
\end{itemize}

\paragraph{Tier 2: Contracted / Striving, Fear, Manipulation}
\begin{itemize}
    \item ``I keep myself busy so I don’t feel the heaviness underneath.''
    \item ``If I can make them doubt their memory, I stay safe.''
    \item ``I sense danger behind their words, attention, and actions, so I play it safe by only saying what they want to hear.''
\end{itemize}

\paragraph{Tier 3: Defensive / Conflict, Ego, Anger}
\begin{itemize}
    \item ``I hate seeing his name pop up; it instantly ruins my mood.''
    \item ``I feel attacked easily and respond by pushing back.''
    \item ``Ugh, I don't mix with people beneath me.''
\end{itemize}

\paragraph{Tier 4: Activated / Transitional, Accepting, Courage}
\begin{itemize}
    \item ``Whatever comes, I'll face it one step at a time.''
    \item ``I'm not waiting for permission anymore; I start to trust my voice.''
    \item ``I accept what has already happened and focus on what I can do next.''
\end{itemize}

\paragraph{Tier 5: Constructive / Growth-Oriented, Relaxed, Healing}
\begin{itemize}
    \item ``I keep my word, even when it's hard.''
    \item ``I want to give because I’m becoming someone who can give, and what I share can actually make a difference.''
    \item ``No matter how busy I am, I take breaks to relax my eyes and body.''
\end{itemize}

\paragraph{Tier 6: Integrated / Harmonious, Clarity, Reasoning}
\begin{itemize}
    \item ``Comprehension feels like light turning on inside my mind.''
    \item ``I can see the pattern beneath the surface.''
    \item ``Let's examine the possible causes of the issue piece by piece.''
\end{itemize}

\paragraph{Tier 7: Expansive / Unifying, Grace, Wisdom}
\begin{itemize}
    \item ``Love doesn't ask why; it simply shines.''
    \item ``Even difficulty feels part of a larger whole.''
    \item ``When the self dissolves, nothing is abandoned and nothing is gained.''
\end{itemize}

\paragraph{Note.}
These examples are provided for reference only. All analyses reported in the main paper are based exclusively on the annotated dataset described in the Dataset Section.

\section{Confusion Matrices for Tier Classification Across Five Representations}
\label{app:confusion_mat_all}

\begin{figure}[!htbp]
\centering
\begin{subfigure}[t]{0.48\linewidth}
    \centering
    \includegraphics[width=\linewidth]{confusion_BAAI_bge_large_en_v1_5.png}
    \caption{BGE-large-en-v1.5}
\end{subfigure}
\hfill
\begin{subfigure}[t]{0.48\linewidth}
    \centering
    \includegraphics[width=\linewidth]{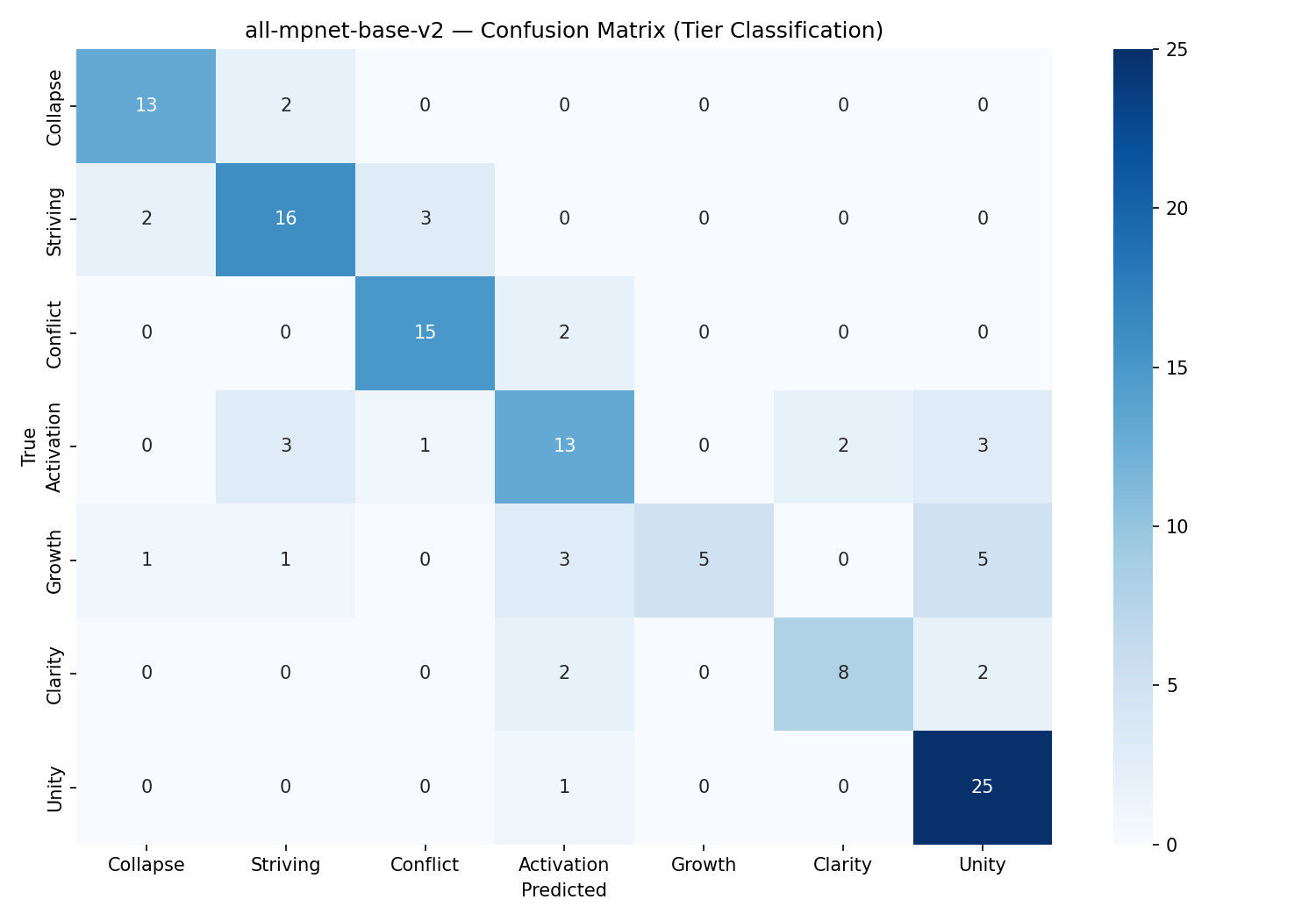}
    \caption{MPNet-base-v2}
\end{subfigure}

\vspace{0.5em}
\begin{subfigure}[t]{0.48\linewidth}
    \centering
    \includegraphics[width=\linewidth]{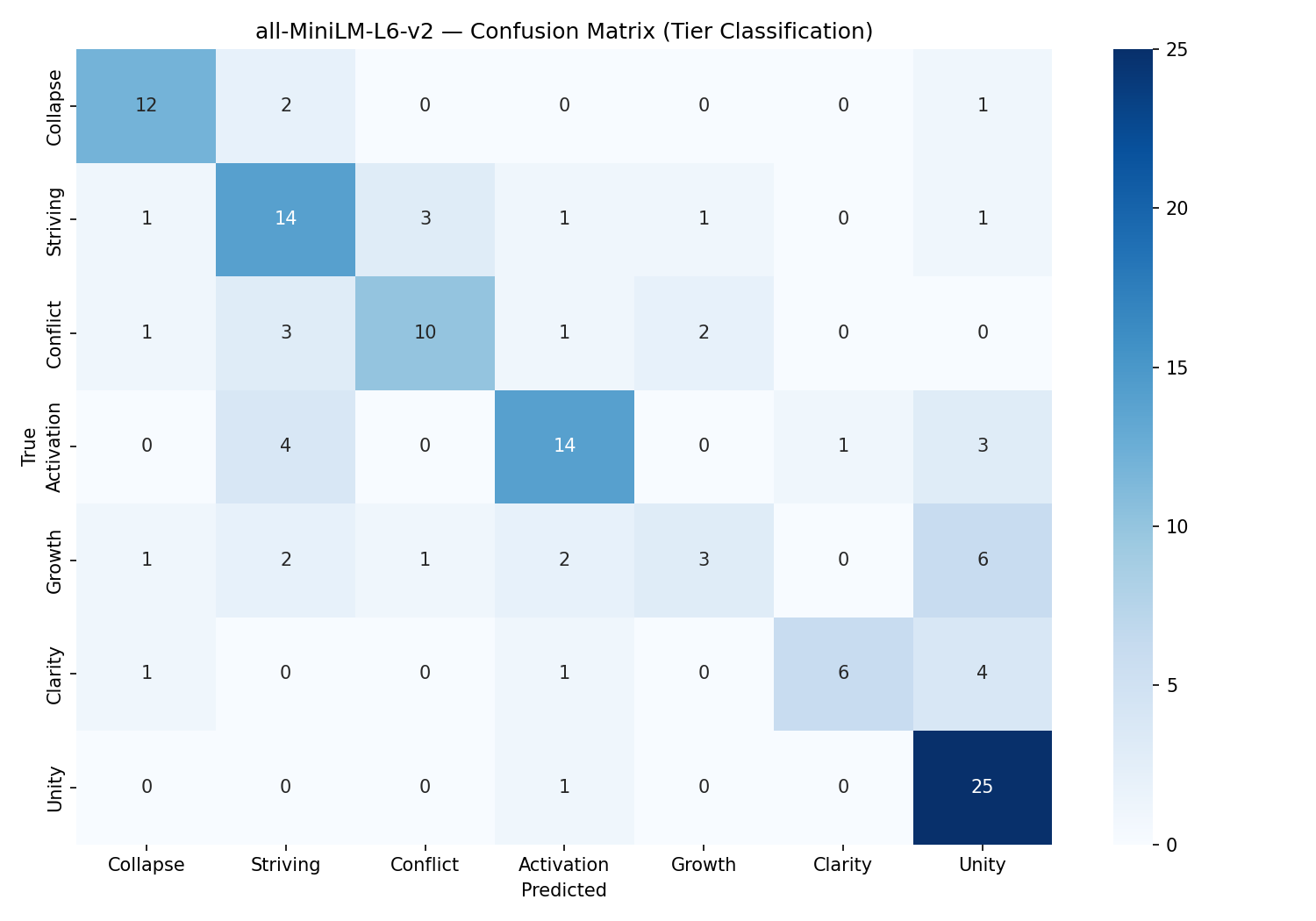}
    \caption{MiniLM-L6-v2}
\end{subfigure}
\hfill
\begin{subfigure}[t]{0.48\linewidth}
    \centering
    \includegraphics[width=\linewidth]{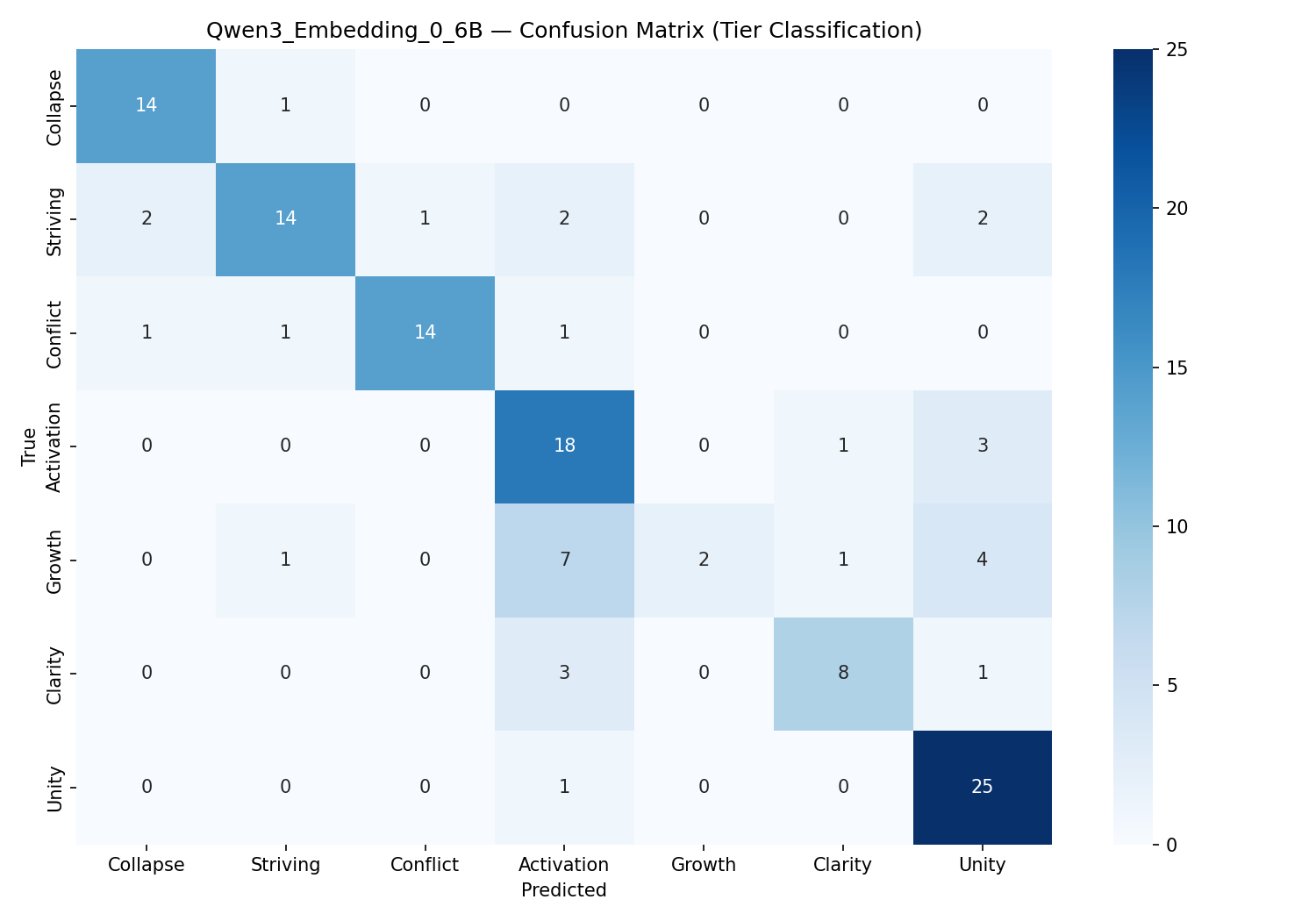}
    \caption{Qwen3-Embedding-0.6B}
\end{subfigure}

\vspace{0.5em}
\begin{subfigure}[t]{0.48\linewidth}
    \centering
    \includegraphics[width=\linewidth]{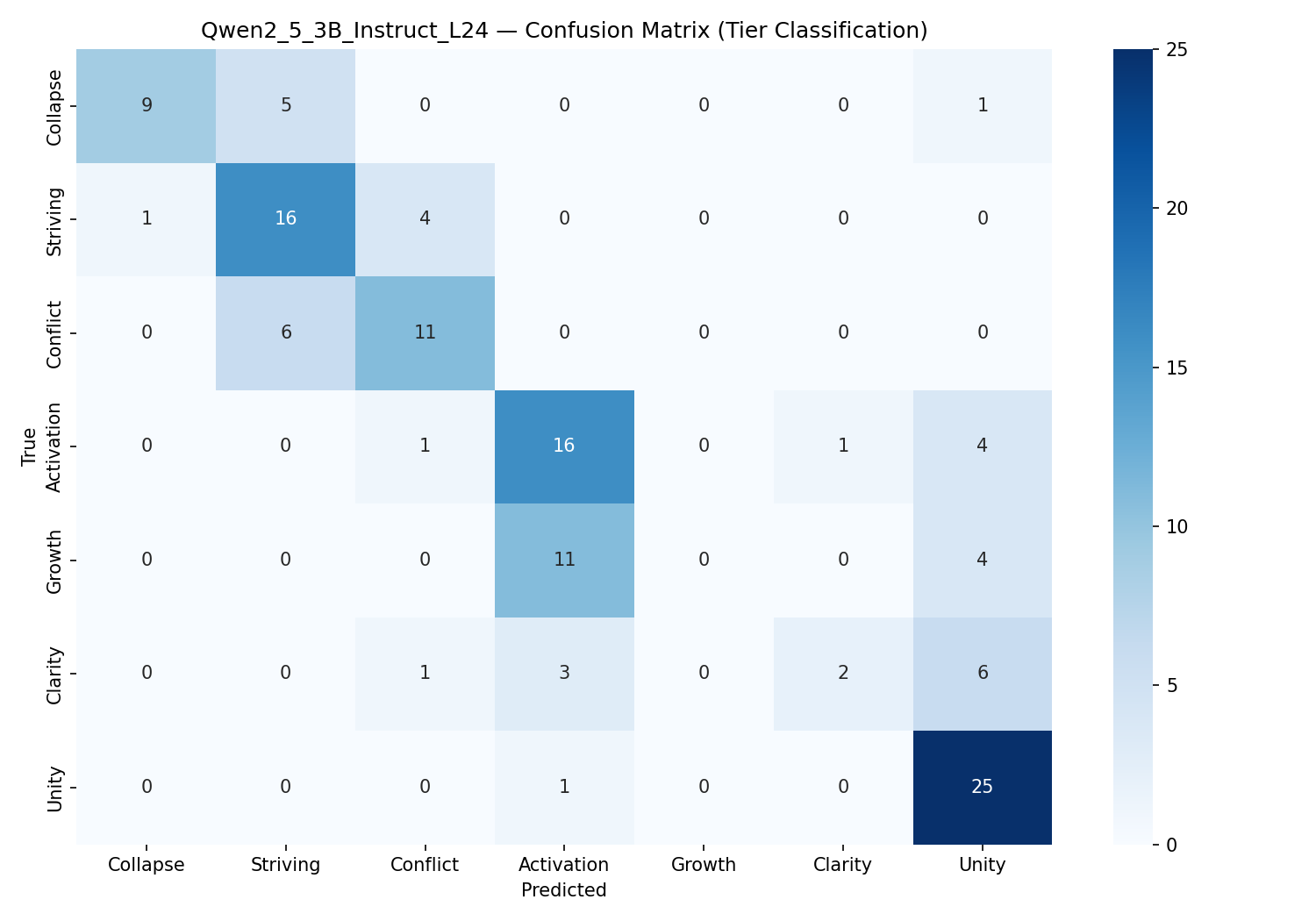}
    \caption{Qwen2.5-3B-Instruct L24}
\end{subfigure}
\caption{Confusion matrices for linear tier classification across five transformer representations.}
\label{fig:confusion_m}
\end{figure}

This appendix presents confusion matrices for tier classification across all five evaluated representations (Figure~\ref{fig:confusion_m}), illustrating that misclassifications occur predominantly between adjacent tiers rather than distant ones.

\section{2D UMAP Visualization of Continuous Energy Scores}
\label{app:2D_umap_all}

This appendix provides 2D UMAP visualizations of transformer representations colored by continuous energy scores (Figure~\ref{fig:umap_2d}), offering a qualitative view of the graded geometric structure discussed in the main text.

\begin{figure}[!htbp]
\centering
\includegraphics[width=\linewidth]{umap_plot_all5_2d_bge_mpnet_minilm_qwen3_qwen25.png}
\caption{2D UMAP visualization of five transformer representations colored by continuous energy scores.}
\label{fig:umap_2d}
\end{figure}

\section{Monte Carlo Permutation Procedure}
\label{app:montecarlo}

Let $T_{\text{obs}}$ denote the observed test statistic. We approximate the null distribution via Monte Carlo permutation by repeating the following procedure $N=200$ times:
\begin{enumerate}
    \item Randomly permute target labels $y' \leftarrow \pi(y)$ using a fixed random number generator seed.
    \item Apply the same probing protocol as in Section~\ref{sec:embeddings} using the same split seeds.
    \item Compute the mean test statistic $T_i$ across splits.
\end{enumerate}

A one-sided permutation $p$-value is computed using the smoothed estimator \citep{good2013permutation}:
\[
p = \frac{1 + \sum_{i=1}^{N} \mathbb{I}[T_i \ge T_{\text{obs}}]}{N + 1}.
\]

\section{Additional Permutation Test Results}
\label{app:permutation_test_extra_his}

This appendix presents permutation test histograms for all five evaluated representations. Each pair of panels shows the shuffled-label null distribution for score regression and tier classification, with the vertical line indicating the observed probe performance.

\begin{figure}[H]
\centering
\begin{subfigure}[t]{0.48\linewidth}
    \centering
    \includegraphics[width=\linewidth]{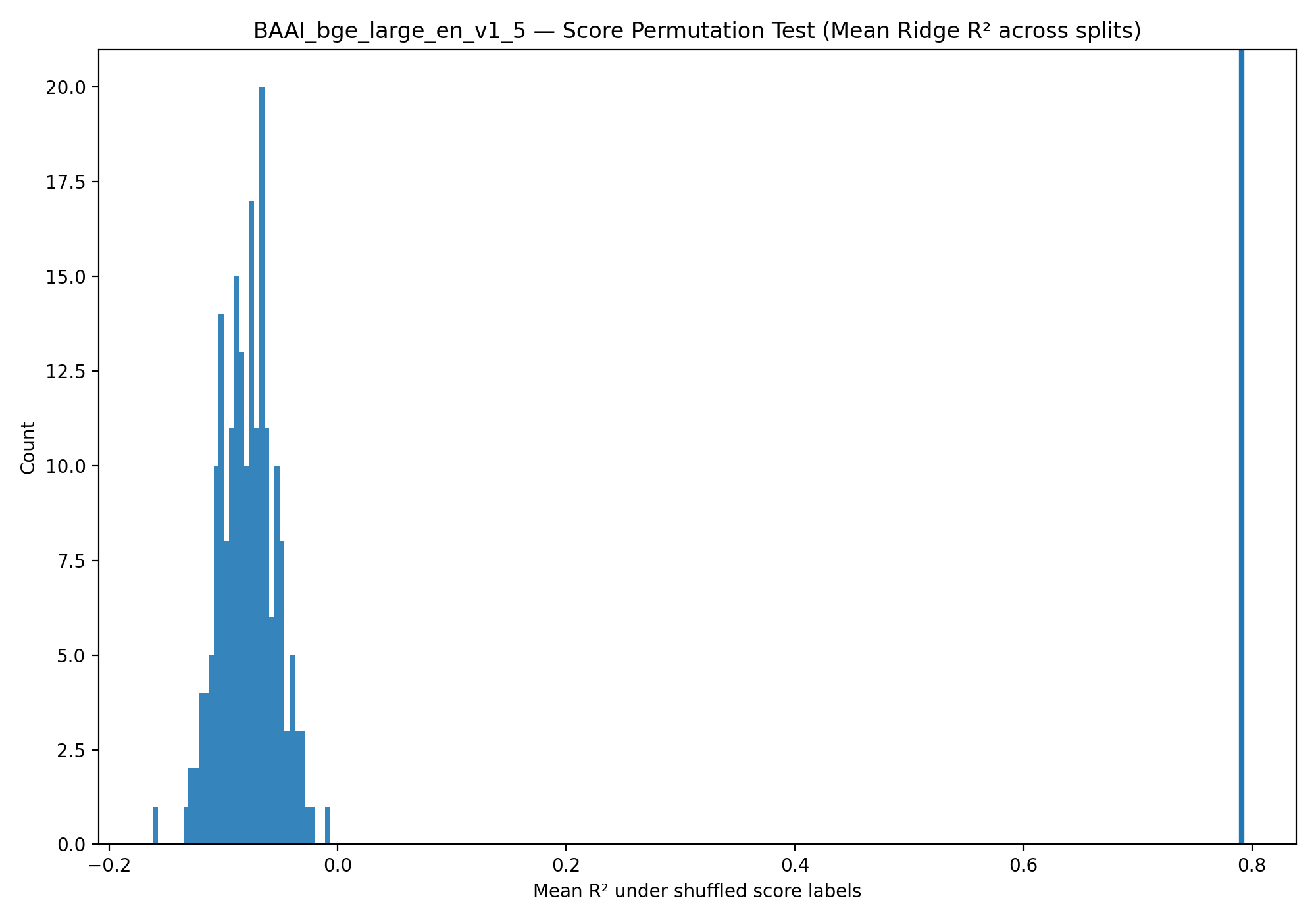}
    \caption{Score permutation test (Ridge $R^2$)}
\end{subfigure}
\hfill
\begin{subfigure}[t]{0.48\linewidth}
    \centering
    \includegraphics[width=\linewidth]{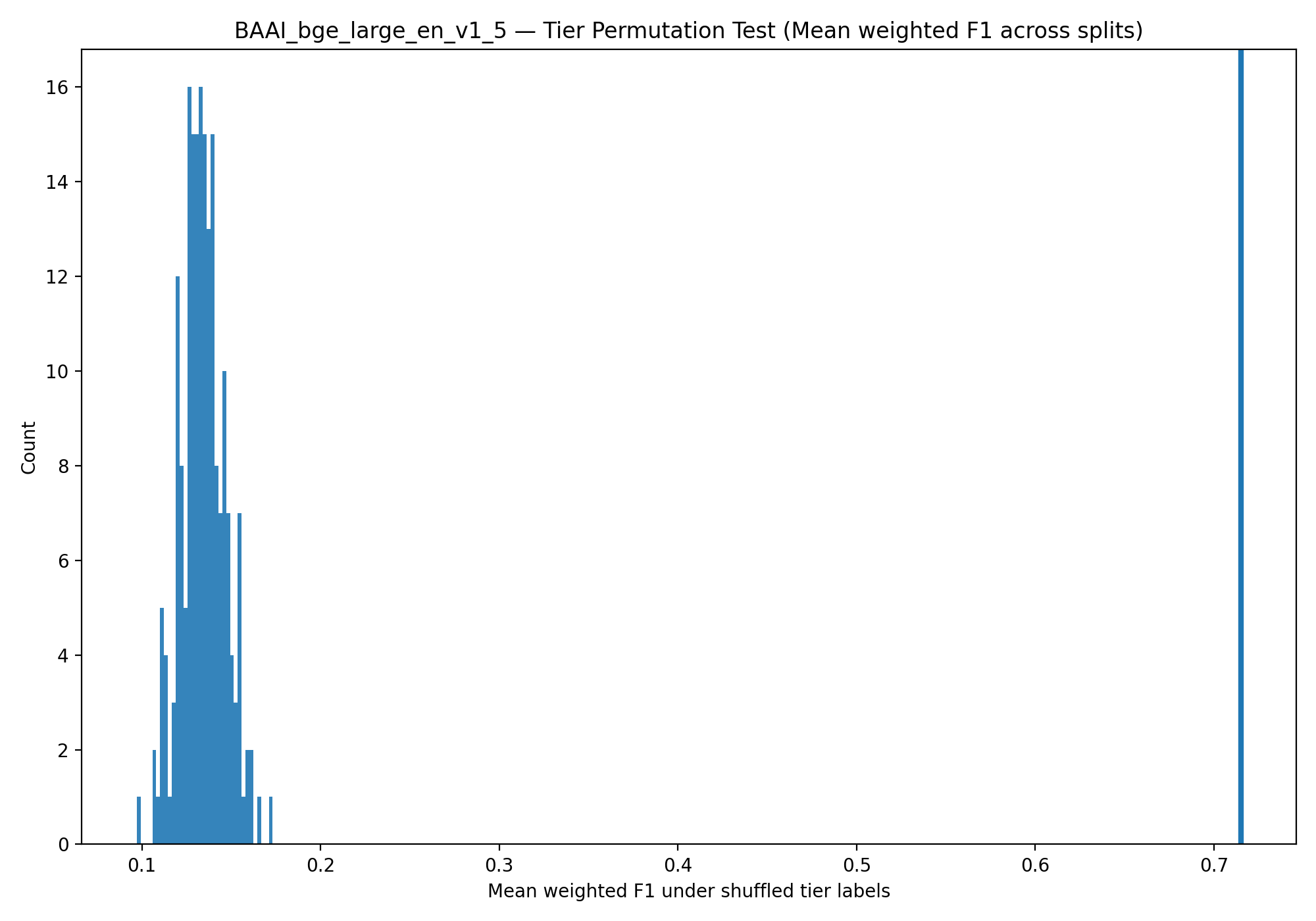}
    \caption{Tier permutation test (Weighted F1)}
\end{subfigure}
\caption{Permutation test results for score regression and tier classification (BGE-large-en-v1.5). Vertical lines indicate observed probe performance.}
\label{fig:perm_test_bge}
\end{figure}

\paragraph{Model: BGE-large-en-v1.5.}
Figure~\ref{fig:perm_test_bge} shows permutation test histograms for score regression and tier classification.

\begin{figure}[H]
\centering
\begin{subfigure}[t]{0.48\linewidth}
    \centering
    \includegraphics[width=\linewidth]{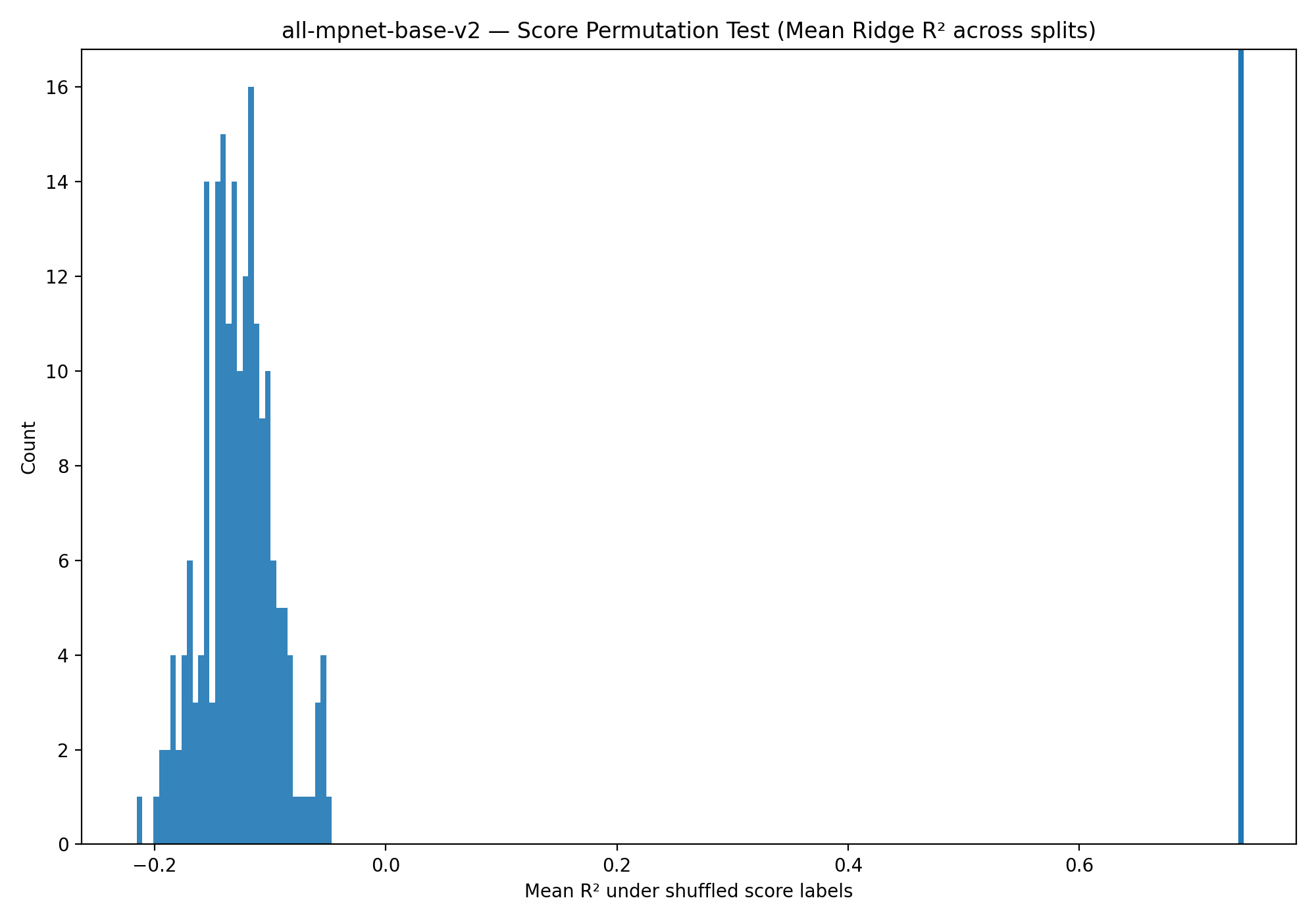}
    \caption{Score permutation test (Ridge $R^2$)}
\end{subfigure}
\hfill
\begin{subfigure}[t]{0.48\linewidth}
    \centering
    \includegraphics[width=\linewidth]{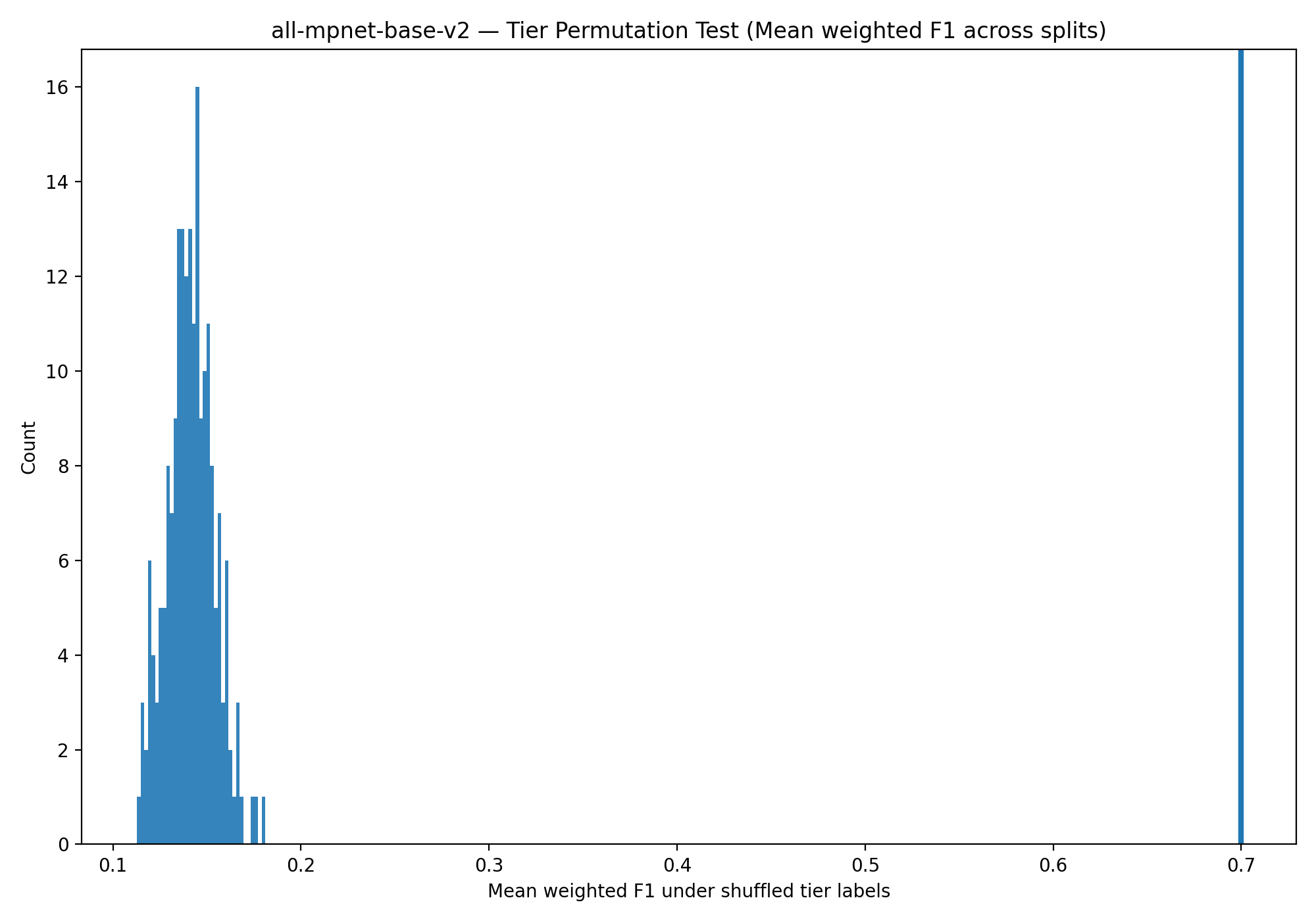}
    \caption{Tier permutation test (Weighted F1)}
\end{subfigure}
\caption{Permutation test results for score regression and tier classification (all-mpnet-base-v2). Vertical lines indicate observed probe performance.}
\label{fig:perm_test_mpnet}
\end{figure}

\paragraph{Model: all-mpnet-base-v2.}
Figure~\ref{fig:perm_test_mpnet} shows permutation test histograms for score regression and tier classification.

\begin{figure}[H]
\centering
\begin{subfigure}[t]{0.48\linewidth}
    \centering
    \includegraphics[width=\linewidth]{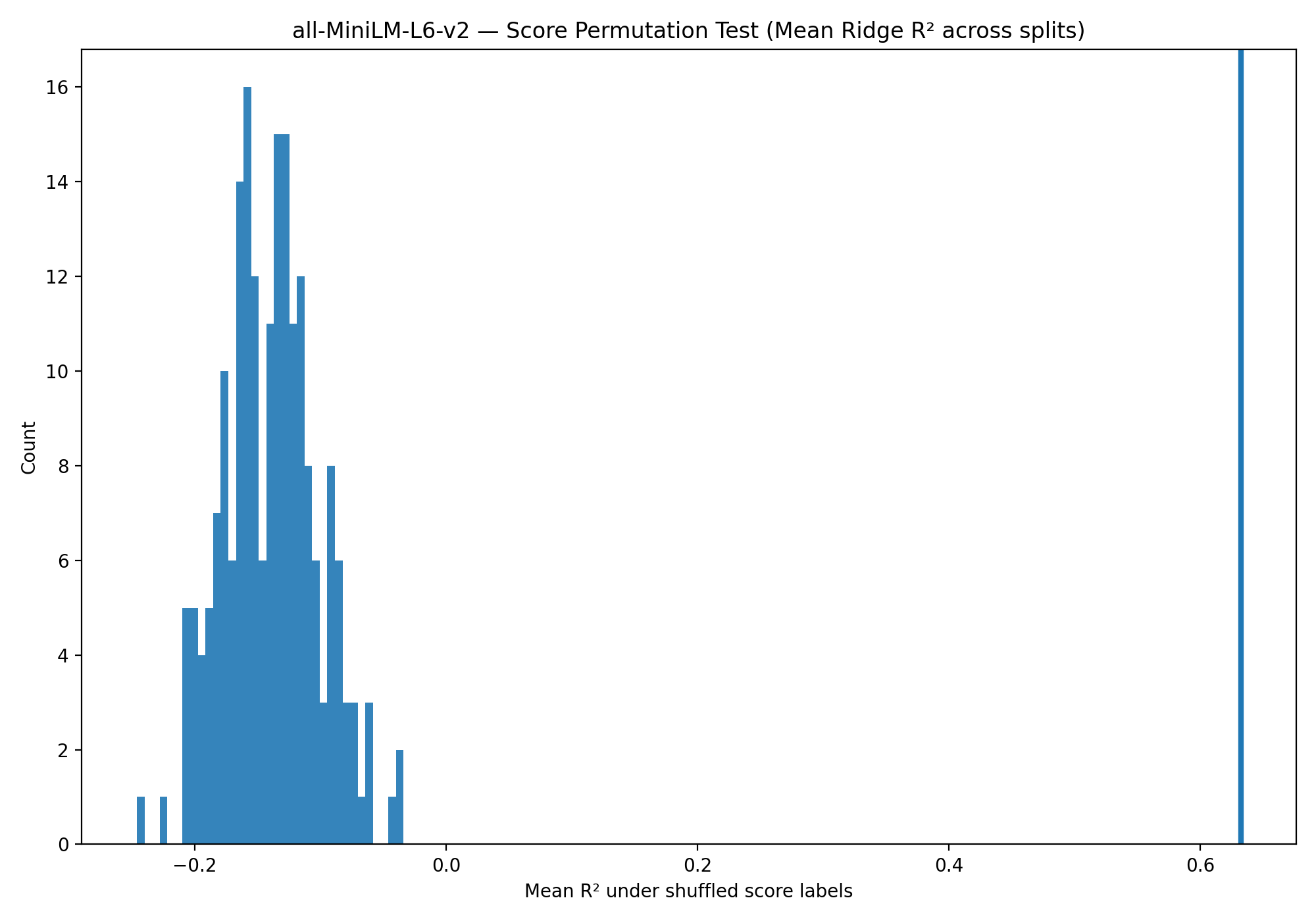}
    \caption{Score permutation test (Ridge $R^2$)}
\end{subfigure}
\hfill
\begin{subfigure}[t]{0.48\linewidth}
    \centering
    \includegraphics[width=\linewidth]{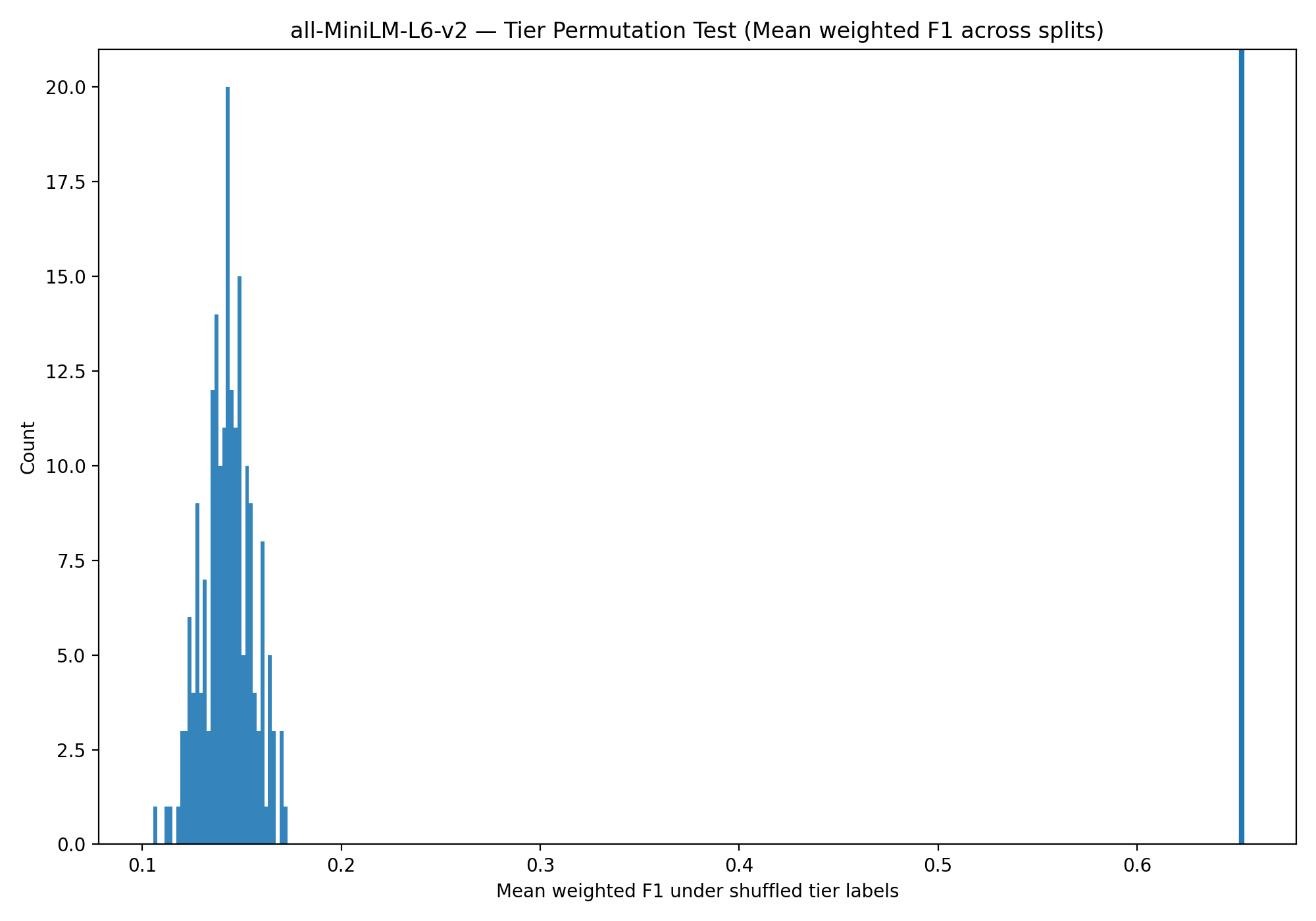}
    \caption{Tier permutation test (Weighted F1)}
\end{subfigure}
\caption{Permutation test results for score regression and tier classification (all-MiniLM-L6-v2). Vertical lines indicate observed probe performance.}
\label{fig:perm_test_minilm}
\end{figure}

\paragraph{Model: all-MiniLM-L6-v2.}
Figure~\ref{fig:perm_test_minilm} shows permutation test histograms for score regression and tier classification.

\begin{figure}[H]
\centering
\begin{subfigure}[t]{0.48\linewidth}
    \centering
    \includegraphics[width=\linewidth]{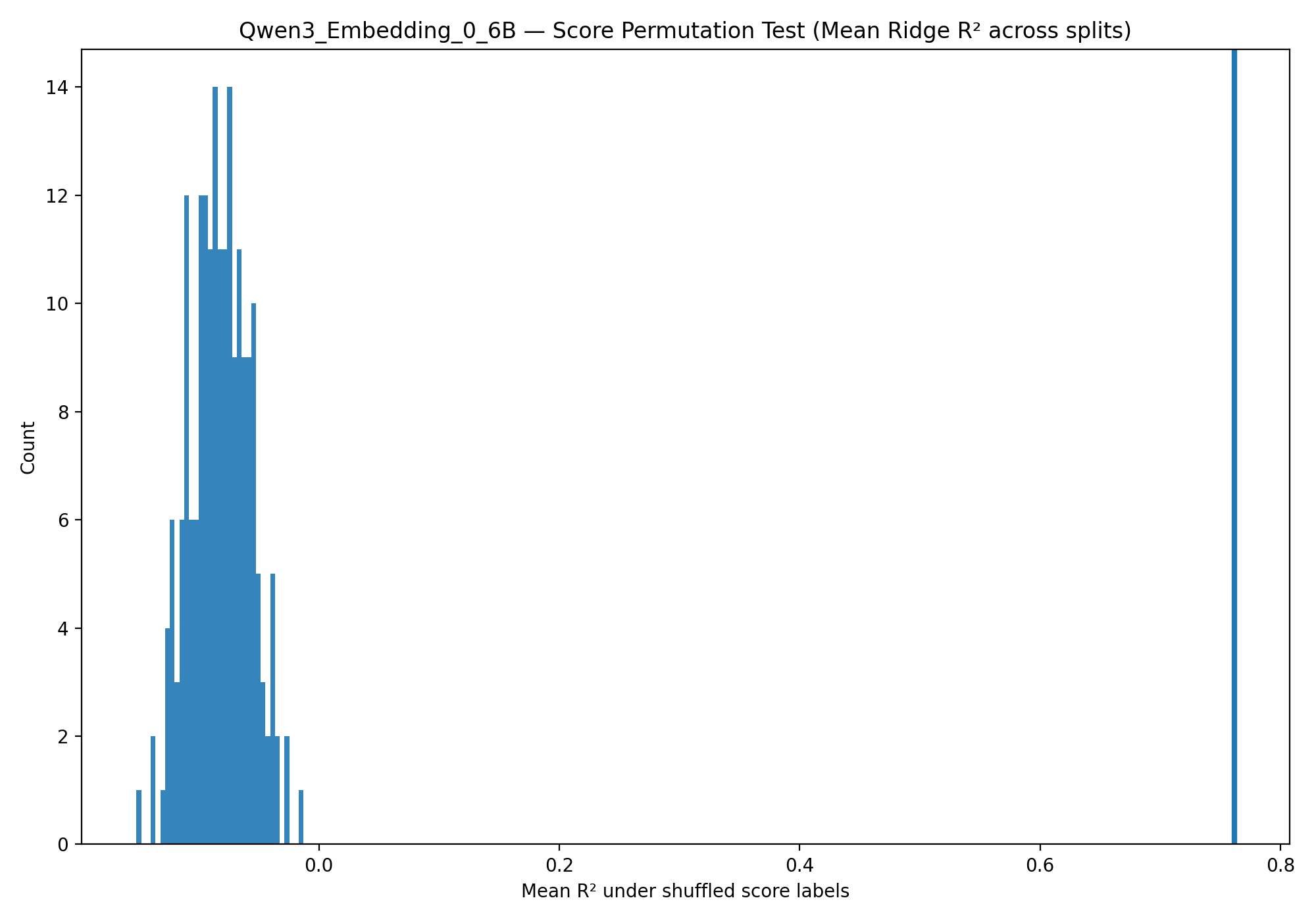}
    \caption{Score permutation test (Ridge $R^2$)}
\end{subfigure}
\hfill
\begin{subfigure}[t]{0.48\linewidth}
    \centering
    \includegraphics[width=\linewidth]{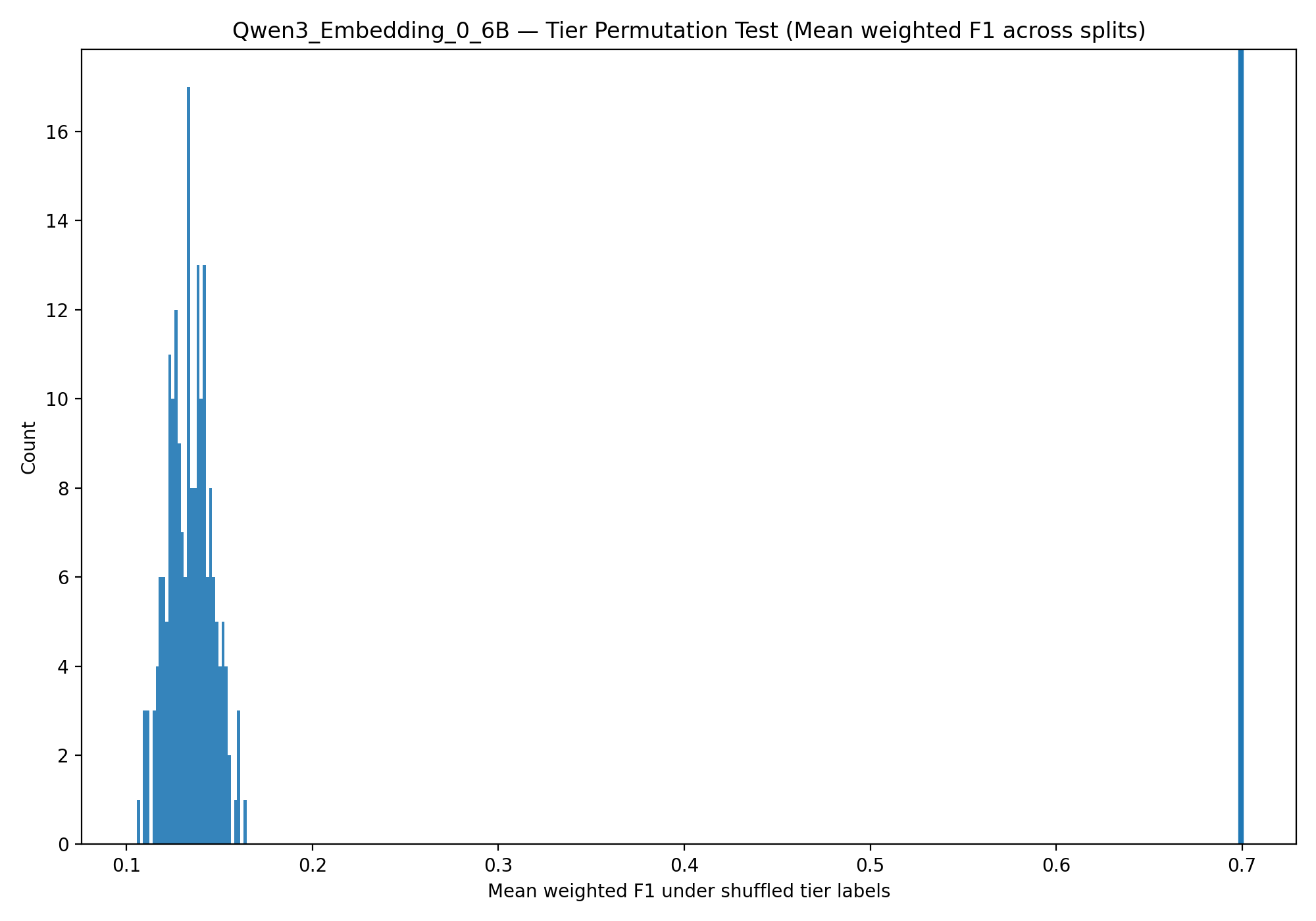}
    \caption{Tier permutation test (Weighted F1)}
\end{subfigure}
\caption{Permutation test results for score regression and tier classification (Qwen3-Embedding-0.6B). Vertical lines indicate observed probe performance.}
\label{fig:perm_test_qwen3}
\end{figure}

\paragraph{Model: Qwen3-Embedding-0.6B.}
Figure~\ref{fig:perm_test_qwen3} shows permutation test histograms for score regression and tier classification.

\begin{figure}[H]
\centering
\begin{subfigure}[t]{0.48\linewidth}
    \centering
    \includegraphics[width=\linewidth]{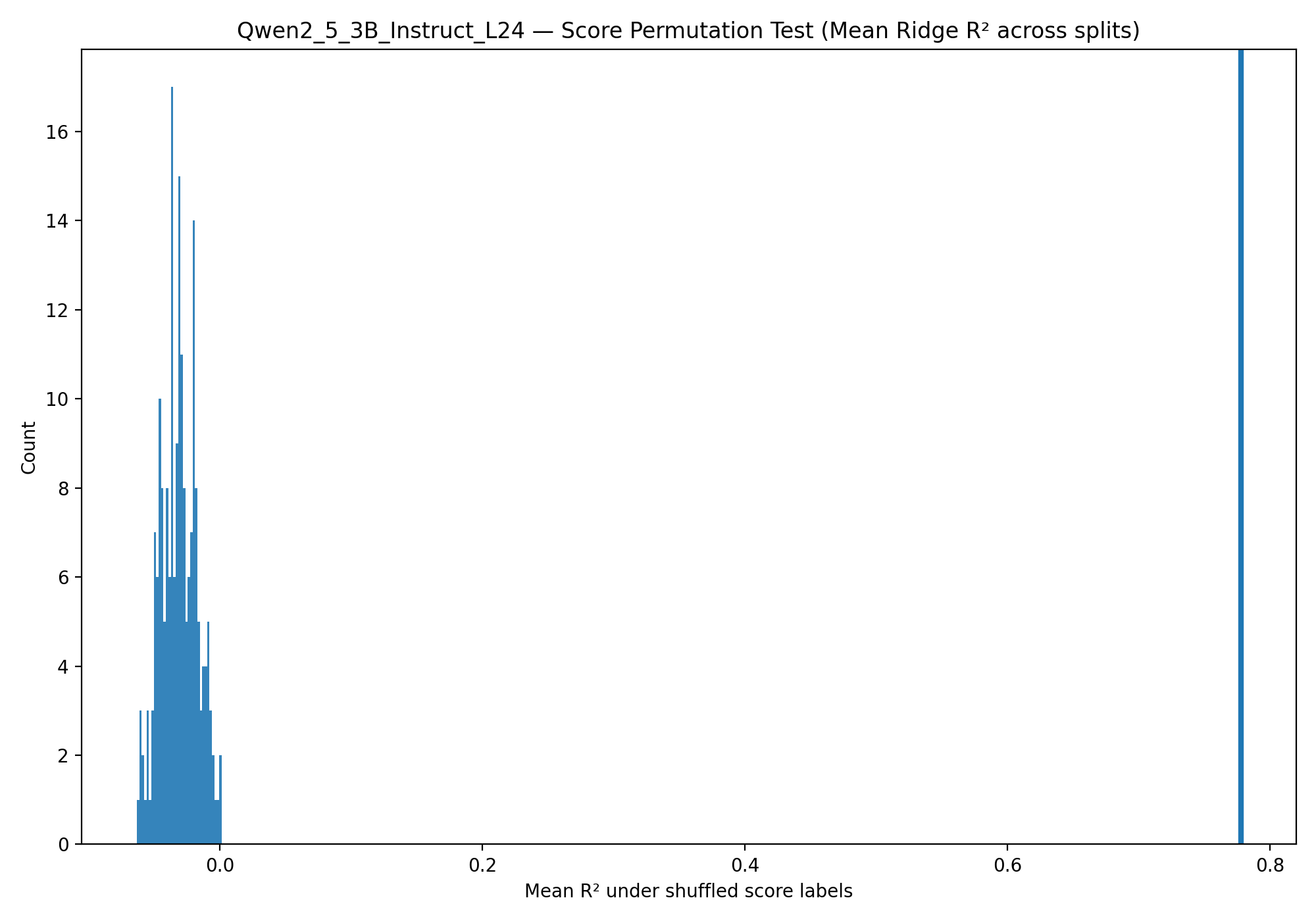}
    \caption{Score permutation test (Ridge $R^2$)}
\end{subfigure}
\hfill
\begin{subfigure}[t]{0.48\linewidth}
    \centering
    \includegraphics[width=\linewidth]{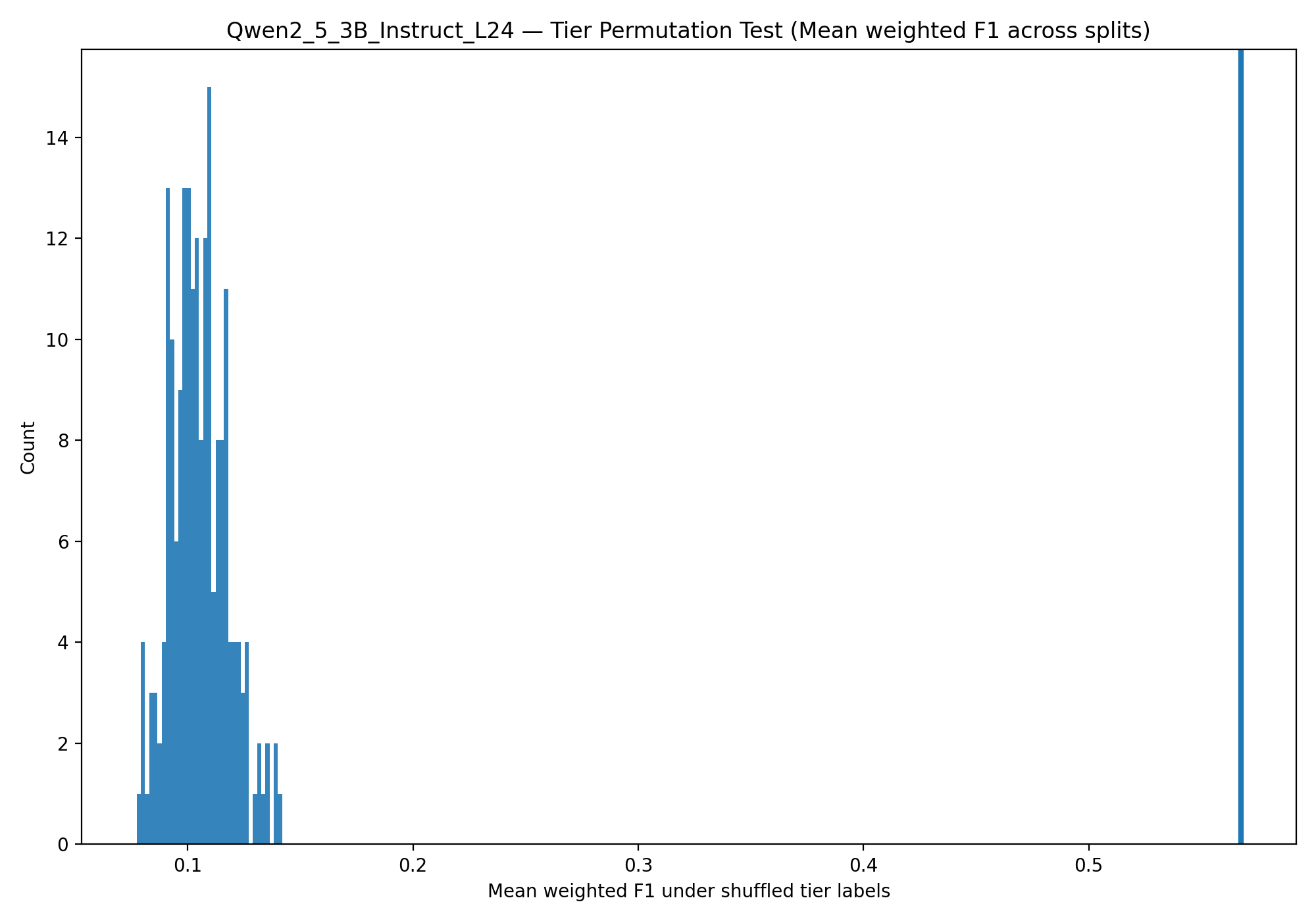}
    \caption{Tier permutation test (Weighted F1)}
\end{subfigure}
\caption{Permutation test results for score regression and tier classification (Qwen2.5-3B-Instruct L24). Vertical lines indicate observed probe performance.}
\label{fig:perm_test_qwen25}
\end{figure}

\paragraph{Model: Qwen2.5-3B-Instruct L24.}
Figure~\ref{fig:perm_test_qwen25} shows permutation test histograms for score regression and tier classification.

\section{Additional Directional Ablation Analyses}
\label{app:directional_ablation}

This appendix presents additional analyses supporting the directional ablation results reported in Section~\ref{sec:directional_ablation_results}. Specifically, we examine (i)  the geometric prominence of the learned direction, (ii) control ablations using randomly sampled directions, and (iii) control ablations using permuted-label directions.

\subsection{Effect on Tier Classification}

A comparable but smaller decrease is observed for tier classification
(Table~\ref{tab:tier_classification_compare}). This pattern indicates
that the ablated direction contributes not only to continuous score prediction
but also to the broader organization of state-of-mind annotations within embedding
space. Notably, the reduction is less pronounced than in the regression task,
suggesting that classification decisions rely on additional geometric features
beyond a single dominant direction.
\begin{table}[!htbp]
\centering
\small
\caption{Tier classification performance before and after directional ablation,
averaged over 20 train--test splits.}
\label{tab:tier_classification_compare}
\begin{tabular}{lcccc}
\hline
& \multicolumn{2}{c}{Logistic Regression} & \multicolumn{2}{c}{MLP (128,64)} \\
Model & F1 (Orig.) $\uparrow$ & F1 (Ablated) $\uparrow$ & F1 (Orig.) $\uparrow$ & F1 (Ablated) $\uparrow$ \\
\hline
BGE-large-en-v1.5         & 0.712 & 0.673 & 0.715 & 0.682 \\
MPNet-base-v2              & 0.705 & 0.655 & 0.705 & 0.659 \\
MiniLM-L6-v2                & 0.652 & 0.596 & 0.635 & 0.564 \\
Qwen3-Embedding-0.6B      & 0.704 & 0.659 & 0.727 & 0.678 \\
Qwen2.5-3B-Instruct (L24) & 0.568 & 0.499 & 0.675 & 0.623 \\
\hline
\end{tabular}
\end{table}

\subsection{Geometric Prominence of the Learned Direction}

To quantify how strongly the learned direction contributes to the overall representation geometry, we measure the variance captured along $\hat{w}$ relative to the total representation variance, computed as the trace of the covariance matrix. Formally, we evaluate
\[
\text{Ratio} = \frac{\mathrm{Var}(X_c \hat{w})}{\mathrm{trace}(\mathrm{Cov}(X_c))},
\]
where $X_c$ denotes mean-centered representations.

As shown in Table~\ref{tab:variance_share}, the learned direction accounts for 1.63\%--4.82\% of total variance across representations. Although modest in absolute magnitude, this variance share is substantially larger than that of random directions sampled in the same space. In all five representations, the learned score direction exceeds all 500 randomly sampled directions, placing it at the empirical 100th percentile among sampled random directions.

\begin{table}[!htbp]
\centering
\small
\caption{Variance share of the learned score-aligned direction relative to total representation variance.}
\label{tab:variance_share}
\begin{tabular}{lccc}
\toprule
Representation & Var(direction) & Trace(Cov) & Ratio (\%) \\
\midrule
BGE-large-en-v1.5 & 0.0179 & 0.4981 & 3.59 \\
MPNet-base-v2 & 0.0169 & 0.8157 & 2.07 \\
MiniLM-L6-v2 & 0.0134 & 0.8229 & 1.63 \\
Qwen3-Embedding-0.6B & 0.0156 & 0.5310 & 2.94 \\
Qwen2.5-3B-Instruct L24 & 0.0087 & 0.1805 & 4.82 \\
\bottomrule
\end{tabular}
\end{table}

\subsection{Control Ablation by Removing Random Directions}

Across the five representations, the learned score direction captures 1.63\%--4.82\% of total variance, while mean random-direction variance shares range from 0.049\% to 0.259\%. In all cases, the learned direction exceeds all 500 sampled random directions.

\subsection{Control Ablation Using Permuted Labels}

To verify that the observed effects are not simply caused by removing any linear direction, we repeat the ablation procedure using a direction learned from permuted energy labels. Under this control condition, probe performance remains largely unchanged compared to the original non-ablated representations, indicating that the regression collapse observed in Section~\ref{sec:directional_ablation_results} is more closely associated with the annotation-aligned direction rather than a generic dimensionality reduction effect.

\section{Additional Analyses for the TF--IDF Baseline}
\label{app:tfidf}

We performed additional analyses using a TF--IDF baseline. These analyses mirror the evaluation protocol used for transformer representations but are reported here in detail for completeness.

\subsection{Experimental Setup}

Sentences are represented using a TF--IDF vectorizer with unigram and bigram features, $\text{ngram\_range}=(1,2)$, $\text{min\_df}=2$, $\text{max\_df}=0.95$, and sublinear term-frequency scaling. All evaluation procedures follow the same protocol as described in Section~\ref{sec:embeddings}, including 20 repeated 80/20 train--test splits.

\subsection{Energy Regression and Tier Classification}

Table~\ref{tab:tfidf_extended_metrics} reports detailed regression and classification metrics for the TF--IDF baseline. Compared to transformer representations, TF--IDF representations yield substantially lower performance across all tasks, indicating that lexical frequency patterns alone do not capture the structured organization observed in representation space.

\begin{table}[!htbp]
\centering
\small
\caption{TF--IDF baseline performance averaged over 20 train--test splits.}
\label{tab:tfidf_extended_metrics}
\begin{tabular}{lcc}
\toprule
Metric & Linear Probe & MLP Probe \\
\midrule
Energy regression ($R^2$) $\uparrow$ & 0.444 & 0.457 \\
Energy regression (MSE) $\downarrow$ & 5.296 & 5.165 \\
Tier classification (Accuracy) $\uparrow$ & 0.473 & 0.487 \\
Tier classification (Weighted F1) $\uparrow$ & 0.421 & 0.446 \\
\bottomrule
\end{tabular}
\end{table}

\subsection{Confusion Matrix Analysis}

\begin{figure}[!htbp]
\centering
\includegraphics[width=0.5\linewidth]{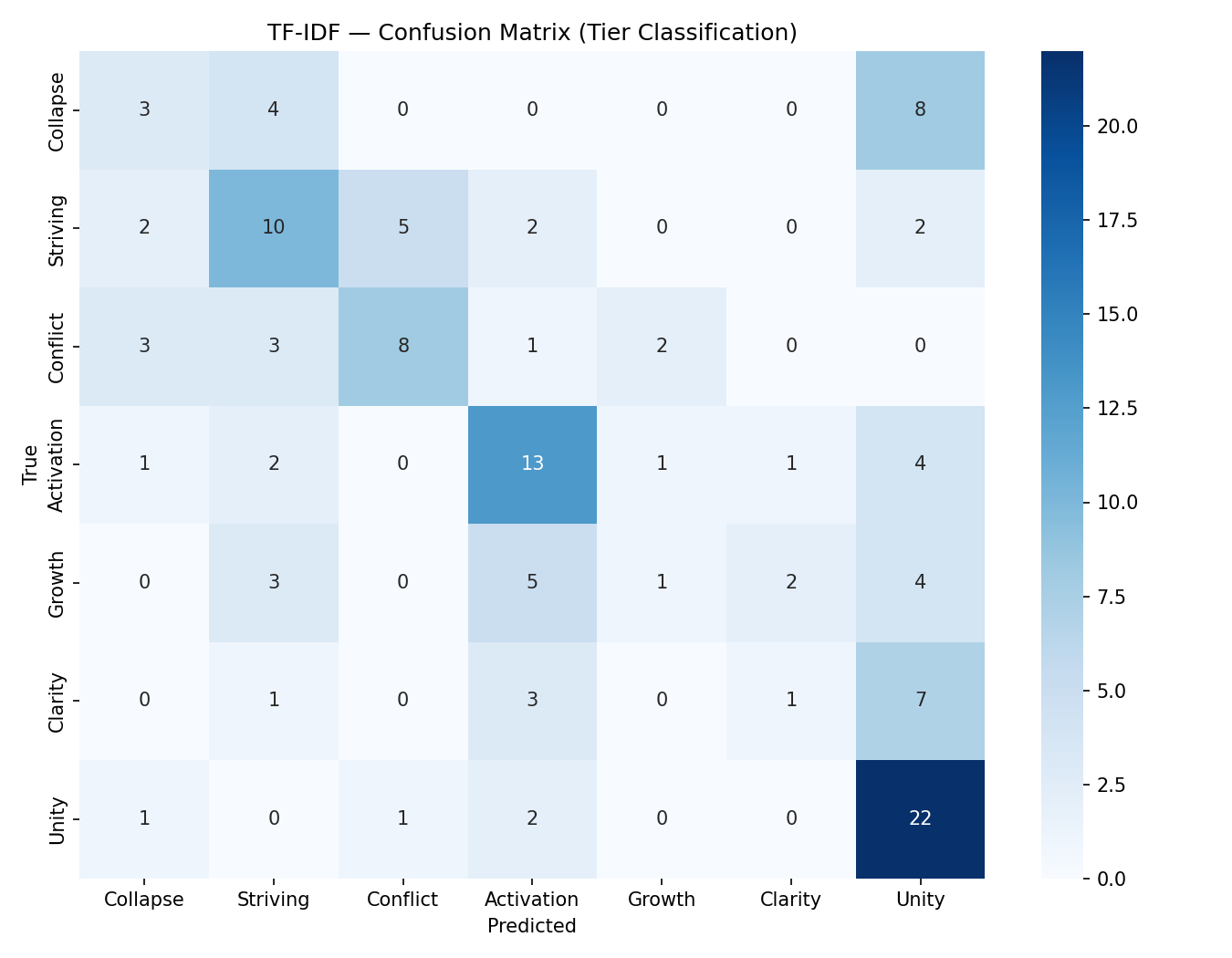}
\caption{TF--IDF linear tier classification confusion matrix.}
\label{fig:tfidf_confusion}
\end{figure}

Figure~\ref{fig:tfidf_confusion} shows the tier classification confusion matrix for the TF--IDF baseline (seed 0). Compared to transformer representations, the TF--IDF representation exhibits weaker separation between adjacent tiers and more diffuse misclassification patterns, suggesting reduced structural coherence.

\subsection{UMAP Visualization}

\begin{figure}[!htbp]
\centering
\begin{subfigure}[t]{0.48\linewidth}
\centering
\includegraphics[width=\linewidth]{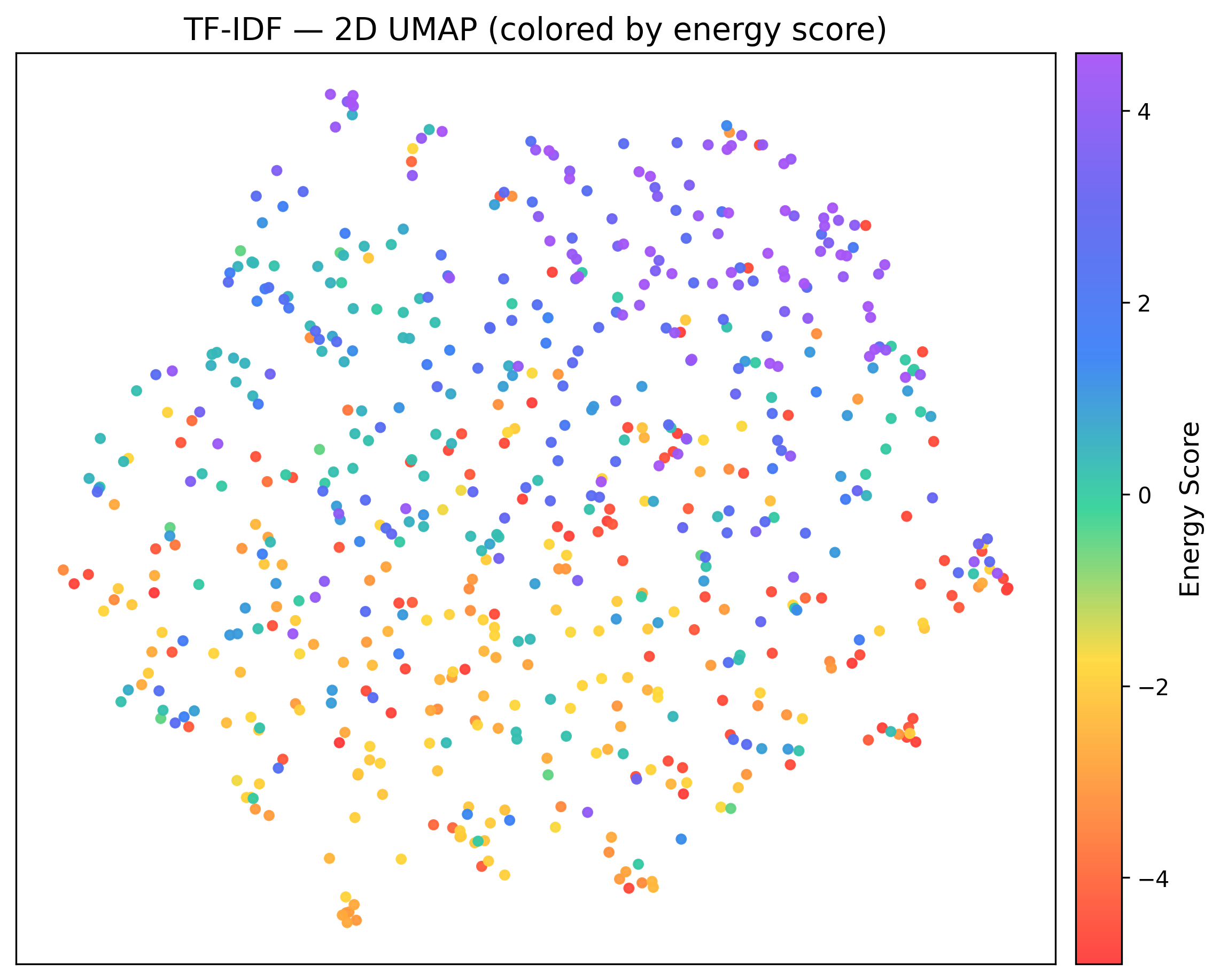}
\caption{2D UMAP}
\label{fig:tfidf_umap2d}
\end{subfigure}
\hfill
\begin{subfigure}[t]{0.48\linewidth}
\centering
\includegraphics[width=\linewidth]{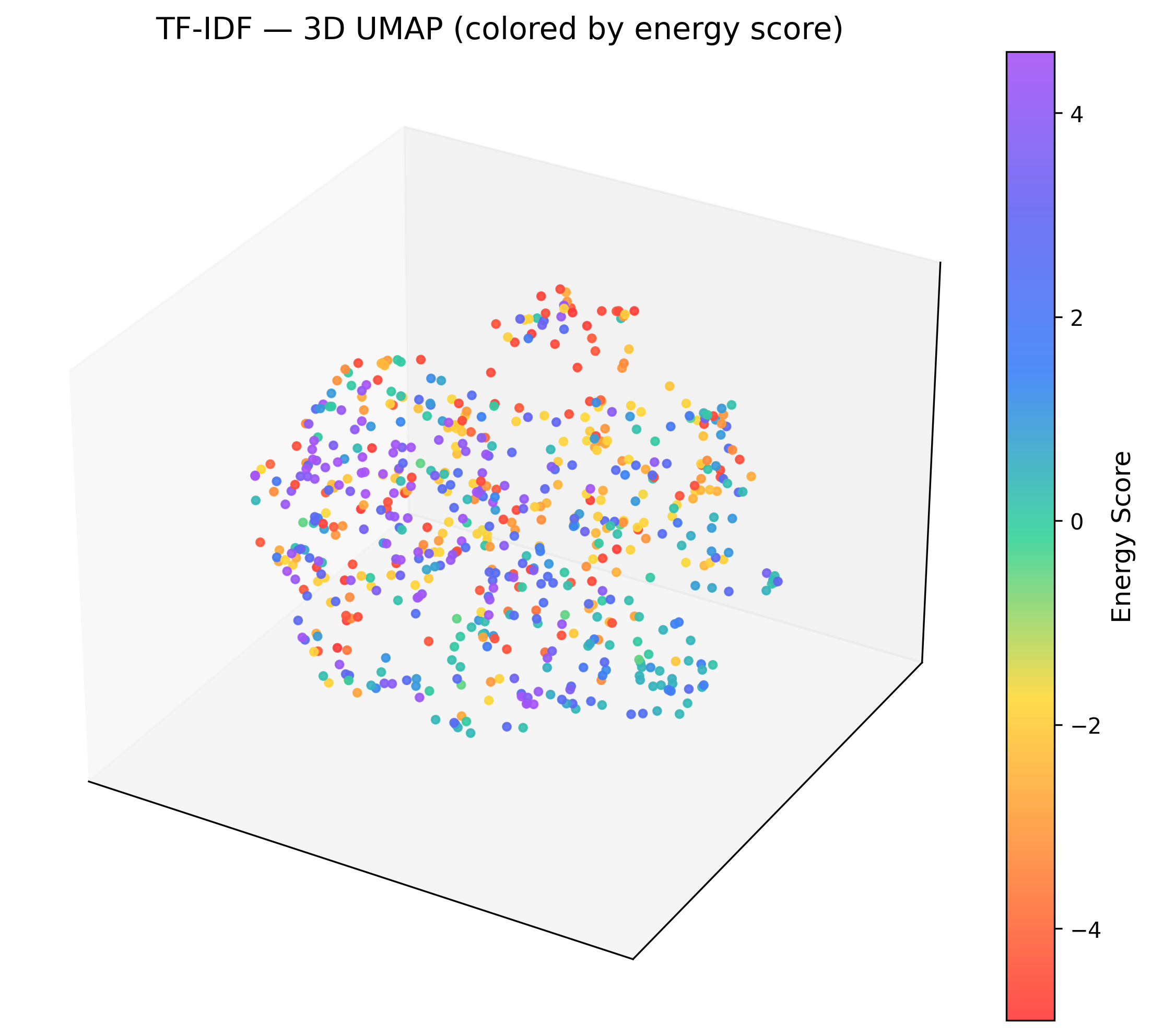}
\caption{3D UMAP}
\label{fig:tfidf_umap3d}
\end{subfigure}
\caption{UMAP visualizations of TF--IDF representations colored by energy score.}
\label{fig:tfidf_umap_combined}
\end{figure}

To provide a qualitative comparison with transformer representations, we visualize TF--IDF representations using 2D and 3D UMAP projections (Figures~\ref{fig:tfidf_umap2d} and~\ref{fig:tfidf_umap3d}). Compared to transformer representations, TF--IDF representations exhibit weaker low-to-high energy gradients and less coherent geometric organization across tiers, indicating that the structure observed in contextual models is not explained by lexical frequency features alone.

Notably, TF--IDF projections still display a faint energy gradient, suggesting that lexical usage partially correlates with the annotated scores. This provides empirical support for the internal linguistic consistency of the annotation scheme, while the substantially stronger organization observed in transformer representations indicates that additional structure is captured beyond lexical statistics.

\end{document}